\documentclass[journal]{IEEEtran}
\usepackage{amsmath,amsfonts}
\usepackage{algorithmic}
\usepackage{array}
\usepackage[caption=false,font=normalsize,labelfont=sf,textfont=sf]{subfig}
\usepackage{textcomp}
\usepackage{stfloats}
\usepackage{url}
\usepackage{verbatim}
\usepackage{graphicx}
\usepackage{lineno}
\def\BibTeX{{\rm B\kern-.05em{\sc i\kern-.025em b}\kern-.08em
    T\kern-.1667em\lower.7ex\hbox{E}\kern-.125emX}}
\usepackage{mathtools} 
\usepackage{cite}
\usepackage{siunitx}
\usepackage{xr}
\usepackage[hidelinks]{hyperref}
\DeclareSIUnit\hour{h}
\newcommand{\V}[1]{\boldsymbol{#1}}
\newcommand{\M}[1]{\mathbf{#1}}

\begin{document}
\title{Series-Parallel Integrated Nonlinear Elastic Actuator applied to the lean motion of a bicycle simulator}
\author{Christina Kohler, Michiel Plooij, Nuria Peña-Perez, Arend L. Schwab, Heike Vallery
\thanks{This work has been funded by the province of Fryslân and the German Federal Ministry of Research, Technology and Space (BMFTR) under the Alexander von Humboldt
foundation and conceptually supported by the BMFTR under the Robotics
Institute Germany (RIG).\\
Christina Kohler and Nuria Peña-Perez are with the Institute of Automatic Control, RWTH Aachen University, 52074 Aachen, Germany. (email: c.kohler@irt.rwth-aachen.de, n.perez@irt.rwth-aachen.de) \\
Michiel Plooij is with Demcon Life Sciences \& Health, 2628 Delft, The Netherlands and with Hapticlink Technologies, 2611 Delft, The Netherlands. (email: michielplooij@hapticlink.nl)\\ 
Arend L. Schwab is with the Department of BioMechanical Engineering,
Delft University of Technology, 2600 Delft, The Netherlands. (email: a.l.schwab@tudelft.nl)\\
Heike Vallery is with the Institute of Automatic Control, RWTH Aachen University, 52074 Aachen, Germany and with the Department of BioMechanical Engineering,
Delft University of Technology, 2600 Delft, The Netherlands, and also with the Department of Rehabilitation Medicine at Erasmus MC, 3015 Rotterdam, The Netherlands. (email: h.vallery@irt.rwth-aachen.de) \\}}

\markboth{Submitted to the IEEE/ASME Transactions on Mechatronics, May 2026}%
{Shell \MakeLowercase{\textit{et al.}}: A Sample Article Using IEEEtran.cls for IEEE Journals}

\maketitle

\begin{abstract}
Designing robots for high-torque, high-fidelity haptic interaction is challenging. Parallel Elastic Actuators (PEAs) use elastic elements in parallel to smaller motors to complement torques, and Series Elastic Actuators (SEAs) use elastic elements in series to decouple motor impedance and improve force control. Recent work combines SEAs and PEAs to obtain both benefits but requires separate elastic elements or clutching.

This paper presents the Series Parallel Integrated Nonlinear Elastic Actuator (SPINEA), which merges SEA and PEA such that a single elastic element takes on dual roles simultaneously, parallel and series. This is achieved by a nonlinear transmission in which the motor and load have misaligned rotation axes and are elastically connected. This geometry enables both high peak torque and precise torque tracking. 

We apply SPINEA to actuate lean of a haptic bicycle simulator, which requires high moments and precise rendering  for safe and realistic rider interactions. We realized a prototype and performed experiments, both with an external excitation setup and with riders cycling. Our results confirm SPINEA's low impedance and precise torque tracking, up to \SI{4.25}{\hertz} with the bicycle frame fixed and up to \SI{4}{\hertz} with riders. The benefits may transfer to other applications requiring compact, high-performance actuation.
\end{abstract}

\begin{IEEEkeywords} Bicycle dynamics, bicycle simulator, compliant actuators, haptics, mechanism design, robotics

\end{IEEEkeywords}

\section{Introduction}
\IEEEPARstart{R}
{ealistic} haptic bicycle simulators could boost cycling training, motor rehabilitation, and traffic safety research. However, accurately emulating the lean dynamics of a bicycle presents an engineering challenge, as it demands both high torque capabilities (to support the gravitational moment of the rider) and excellent torque tracking (to ensure safe and realistic interactions with the rider). 

Conventional cycling simulators typically either exclude the leaning degree of freedom (DoF) \cite{Shoman.2021} or use an actuated platform \cite{Kwon.2001, Wintersberger.2022, MartinezGarcia.2023}. The first approach precludes rendering realistic haptic lean interactions between the rider and the bicycle by design. Actuated platforms allow rotation around the lean axis, but they also require control over high inertia since the platform must support a person's weight. This requires large motors, which can pose safety risks, compromise interaction realism and hinder system accessibility. 

Given their success in other safety-critical applications requiring high forces and precise force control, actuators with elastic elements may offer solutions for actuating the lean of a bicycle simulator, without relying on a moving platform. Two main categories exist: Series and Parallel Elastic Actuators.

Series Elastic Actuators (SEAs) introduce compliance by placing elastic elements in series with the motor, enhancing force control \cite{Pratt.1997, Vallery.2008}. This compliance, which can be variable (as reviewed in \cite{Wolf.2016}), decouples the actuator output from the high inertia of the motor and any mechanical transmissions, allowing for high-fidelity force control. This has been exploited in physical human-robot interaction, for example in rehabilitation applications \cite{Paine.2014, Vallery.2021}. In addition, the series elastic element facilitates force sensing via its elongation \cite{Pratt.1997}. However, a compliant element in series does not augment, but just transmits the motor's torque output. Thus, for SEAs, the motor and transmission system must deliver the full output torque, limiting their use in applications requiring high torque.

Parallel Elastic Actuators (PEAs) place elastic elements in parallel with the actuator, allowing for energy storage and release, which helps generate high torque \cite{Verstraten.2016}. This configuration reduces the motor's workload by sharing it with the elastic elements, allowing for a smaller motor and transmission ratio, as commonly described for PEA designs \cite{Plooij.2015,Krimsky.2024, Liu.2024}. This leads to more responsive actuators with lower output inertia, as highlighted in optimized PEA designs \cite{Liu.2024} and experimentally validated in applications such as torque-controlled exoskeletons \cite{Toxiri.2018}. However, as the parallel compliance in PEAs does not decouple the motor from the load, the inertia and friction of the drivetrain (albeit smaller) remain uncompensated.

Hybrid actuators combine features of both SEAs and PEAs. For example, Grimmer at al.\ showed via simulations in an ankle joint actuator that hybrid actuators can lower motor peak power demands and energy consumption compared to SEAs or PEAs alone \cite{Grimmer.2012}. Physical realizations of hybrid actuators may or may not include a second motor \cite{Roozing.2021, Pfeifer.2015}, but typically assign series and parallel roles to separate elastic elements, leading to complex and less compact mechanical structures. 

Few hybrid actuators allow elastic elements to assume dual roles, either across multiple DoFs~\cite{Wyss.2019} or within a single DoF through clutching~\cite{Mathijssen.2015}. The multidimensional compliant actuator (MUCDA) assigns different roles to springs in different DoFs, acting in series for one and in parallel for others \cite{Wyss.2019}. This is unsuitable for bicycle lean, which requires dual roles within one DoF. The SPEA concept~\cite{Mathijssen.2015} allows springs to switch between series and parallel roles, dynamically adjusting stiffness and torque output. However, existing SPEAs require additional components like dedicated clutches for role switching, as each spring can only fulfil one role at a time.

A potential option to enable dual roles without clutching is nonlinear transmission, benefiting applications that need high torques only in part of the workspace. This applies to bicycle leaning, where torque requirements peak at larger lean angles due to high gravitational moments. As an example of nonlinear transmission in SEAs, our group's  compliant leg prosthesis~\cite{Pfeifer.2015} allowed for a smaller motor by exploiting nonlinear torque and stiffness needs. However, it still required two separate elastic elements for series and parallel functionality.

\begin{figure}
    \centering
    \subfloat{%
        \includegraphics[width=0.69\linewidth]{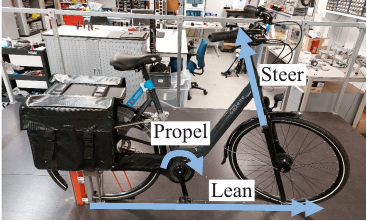}%
    }
    \hfill
    \subfloat{%
        \includegraphics[width=0.27\linewidth]{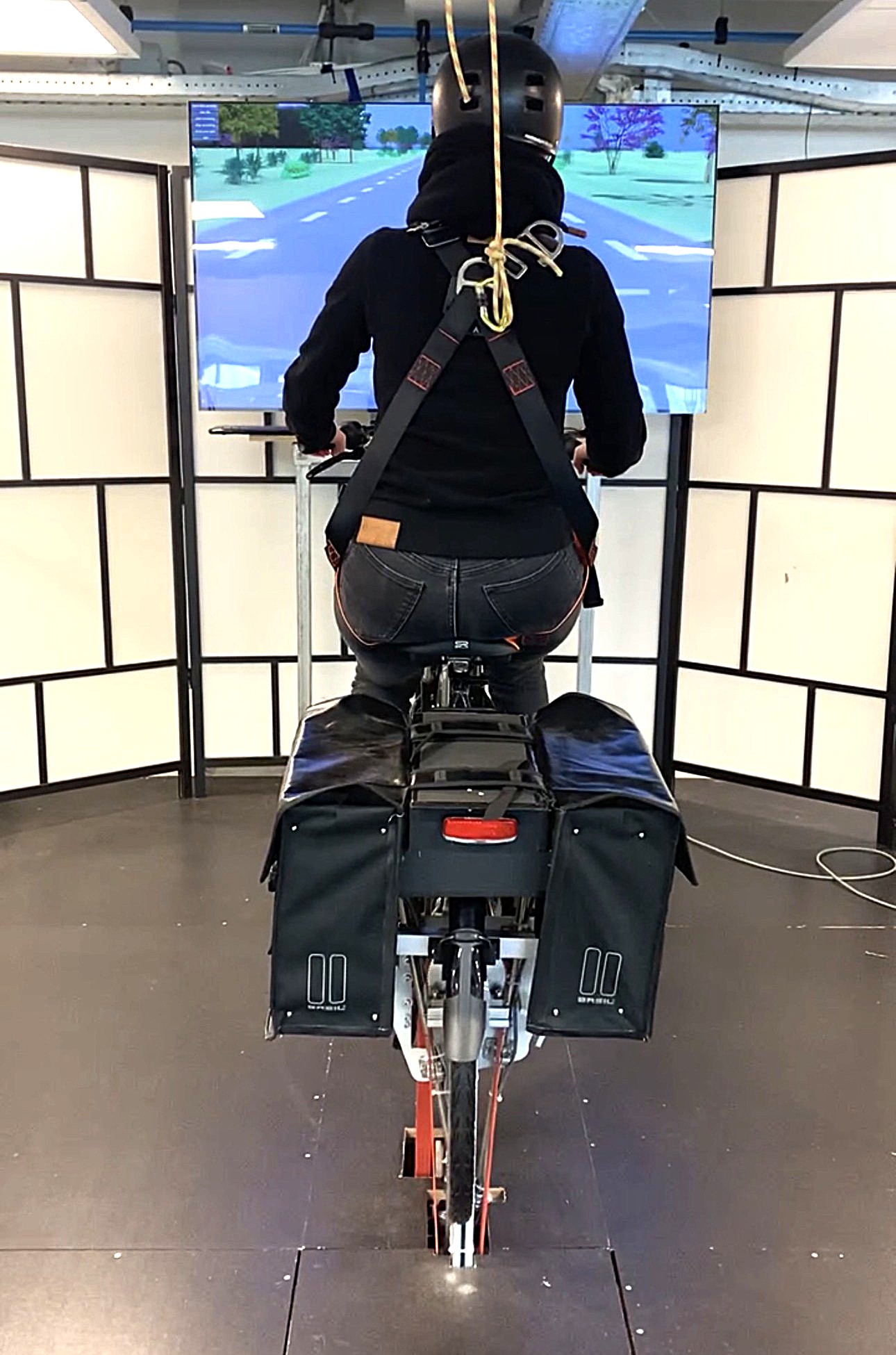}%
    }
    \caption{Bicycle simulator setup at TU Delft. The system looks like a normal bicycle, with most of the actuation and electronics hidden within saddle bags and underneath a raised floor. (a) Side view, showing the three DoFs: lean (enabled by SPINEA), steer, and propel; (b) Back view, with a human rider. Screen displays a virtual environment.} 
    \label{fig:Overview_Simulator}
\end{figure}

This work proposes the Series-Parallel Integrated Nonlinear Elastic Actuator (SPINEA), which combines features of both SEAs and PEAs through a nonlinear arrangement with two antagonistic springs that serve dual roles due to the mechanism's kinematics. No clutching is used to switch roles, leading to a simple design. We implemented SPINEA to actuate the lean motion of a new haptic bicycle simulator (Figs.~\ref{fig:Overview_Simulator}(a), (b)). By leveraging both SEAs and PEAs, SPINEA aims to deliver high-fidelity torques across a wide range of leaning angles during dynamic rider-simulator interactions. Our nonlinear approach addresses high torque requirements by recognizing that gravitational moments peak at the workspace edges, while maintaining actuator \textit{transparency} and simulator \textit{safety}. The transparent actuator ensures that the simulator exhibits only intended (bicycle) dynamics without adding mechanical or control-induced mechanical impedance. Simulator safety is linked to the actuator's ability to passively provide enough torque support against gravitational moments, preventing falls even during power outages.

The rest of this contribution organizes as follows: In Section~\ref{sec:Design}, we derive actuator requirements based on the bicycle simulator application and detail our conceptual design steps, operational concept, and final actuator realization. Section~\ref{sec:Control} presents sensing, signal processing and torque control. Section~\ref{sec:evaluation methodology} describes the experimental evaluation of the actuator's transparency and torque capabilities. We assess transparency by evaluating the torque sensing, impedance characteristics and torque tracking accuracy (including achievable control bandwidth). Section~\ref{sec:results} presents the results and Section~\ref{sec:Discussion} the discussion, with a conclusion in Section~\ref{sec:conclusion}.

\section{SPINEA Hardware Design}
\label{sec:Design}
\subsection{Requirements}
\label{sec:Design_requirements}
We derived the requirements for SPINEA from existing literature on on-road bicycle dynamics and geometry. A summary of the resulting specifications is presented in Table~\ref{Tab:Requirements}.

First, SPINEA must allow a realistic {lean range of motion (RoM)}, while limiting the maximum lean angle for safety. Experimental cycling studies indicate $\pm \SI{2}{\degree}$ lean during straight cycling \cite{Cain.2016} and up to $\pm \SI{20}{\degree}$ \cite{Sanjurjo.2019} and $\pm \SI{30}{\degree}$ \cite{VandenOuden.2011} during steady-state turning. We set the leaning workspace edges at $\pm \SI{20}{\degree}$ to accommodate untrained or vulnerable riders. This choice aims to reduce risks of falls or pedal-ground contact during turning.   

Second, we estimated SPINEA's {torque} requirements using the linearised equations of motion for the lean and steer motion of a bicycle as presented by Meijaard et al. \cite{Meijaard.2007}:
\begin{equation}
	\M{M} \ddot{\V{q}}+v\M{C}_1 \dot{\V{q}} + \left(g \M{K}_0 + v^2 \M{K}_2 \right) \V{q}  = \V{f},
	\label{eq:eom_real}
\end{equation}
with the degrees of freedom $\V{q} = (\varphi,\delta)^T$ containing lean angle $\varphi$ and steer angle $\delta$, the applied generalized torque vector $\V{f}=(\tau_\varphi, \tau_\delta)^T$ containing lean torque $\tau_\varphi$ and steer torque $\tau_\delta$, forward velocity $v$ and gravity $g$. The matrices $\M{M},\M{C}_1, \M{K}_0, \M{K}_2$ are constants that depend on bicycle geometry and mass distribution. The forward velocity $v$ impacts gyroscopic and centrifugal effects as shown by the terms $v\M{C}_1 \dot{\V{q}}$ and $v^2 \M{K}_2 \V{q}$.  

The bicycle simulator consists of an off-the-shelf Dutch city bicycle (Orange+, Gazelle, NL), fixed in the longitudinal direction, but with actuated lean and steer degrees of freedom (Fig.~\ref{fig:Overview_Simulator}(a)). We assume that the simulator shares the same mass and gravitational properties as the bicycle on the street, so that only the velocity-dependent terms from the linearized equations of motion have to be rendered by the lean and steer actuators. Then, the equations of motion for the simulator are:
\begin{equation}
	\M{M} \ddot{\V{q}} + g \M{K}_0 \V{q}  = \V{f} + \V{f}_\mathrm{a}, 
	\label{eq:eom_phys_siml}
\end{equation} 
with the actuator applied torques $\V{f}_\mathrm{a}=(\tau_{\mathrm{a, ref}},\tau_{\delta, \mathrm{ref}})$, the reference lean torque $\tau_{\mathrm{a, ref}}$ and the reference steer torque $\tau_{\delta, \mathrm{ref}}$. The required  actuator torques result from \eqref{eq:eom_real} and \eqref{eq:eom_phys_siml}:
\begin{equation}
\V{f}_\mathrm{a} = -\left(v\M{C}_1 \dot{\V{q}} + v^2 \M{K}_2 \V{q} \right).
\label{eq:controller_straight}
\end{equation}
Details on the specific values for the constant matrices can be found in the Supplementary Materials (Section \ref{sec:Lean_and_steer_torque_actuation_model}).

At workspace edges, high torques are needed to render the velocity-dependent terms, which counterbalance gravity. Assuming a bicycle mass of \SI{21}{\kilo\gram}, a maximum rider mass of \SI{100}{\kilo\gram} and  a combined bicycle-rider centre of mass located \SI{0.9}{\meter} above the ground \cite{Meijaard.2007}, we estimate a maximum gravitational lean torque of \SI{365}{\newton\meter} at a lean angle of \SI{20}{\degree}. The actuator must be capable of delivering at least this torque and hold it passively in the event of power loss. 

Precise haptic rendering is crucial around the upright configuration, where we expect leaning torques to be smaller, to effectively emulate subtle balancing dynamics. The rider should not feel parasitic torques from the dynamics of a powerful actuator, such as inertia or gearbox friction. Instead, the leaning should mimic that of a conventional bicycle. To achieve this, we target low system impedance that resembles the mass properties of an on-road bicycle. Reported {mass moment of inertia} about the lean axis of a riderless bicycle is within \SI{3}- \SI{8}{\kilo\gram\meter\squared} \cite{moore2010accurate, Kooijman.2008} when no battery is installed, and within \SI{8} - \SI{11}{\kilo\gram\meter\squared} \cite{Kalsbeek_2016} when a battery is present. The {first moment of mass} is in the range of $\SI{5} - \SI{11}{\kilo\gram\meter}$ for non-electric bicycles \cite{moore2010accurate, Kalsbeek_2016} and between $\SI{12} - \SI{16}{\kilo\gram\meter}$ \cite{Kooijman.2008} for electric bicycles with batteries.

We specify a {closed-loop bandwidth} of at least \SI{2}{\hertz} to ensure accurate torque tracking that replicates real bicycle dynamics. Moore et al.\cite{Moore.2011} showed that rider steering control inputs mostly remain below \SI{2}{\hertz}. This value aligns with the frequency content observed in lean angle time-series data from other experimental cycling studies \cite{Cain.2016, Sanjurjo.2019, VandenOuden.2011}.

Finally, the simulator must function as both a training and research platform, closely resembling an on-road bicycle. To enable training of mounting and dismounting, the simulator needs a stationary platform to emulate the street, which a priori excludes conventional simulator solutions that rely on moving platforms like hexapods. Additionally, to ensure an {appearance} akin to an on-road bicycle, all mechanical and electrical components must fit underneath the platform or inside standard saddle bags, which measure about \SI{0.45}{\meter} in height and $\SI{0.4}{\meter}$ in width.

\begin{table}
\centering
    \caption{Actuator design requirements}
    \label{Tab:Requirements}
    \begin{tabular}{p{2.7cm} p{1.5cm} p{3cm}}
        \hline
        Requirement & Desired Metric & Rationale \\
        \hline
        Lean RoM   & $\pm \SI{20}{\degree}$   & Ensures safe and realistic leaning motion \\
        Peak torque   & $\pm \SI{365}{\newton\meter}$ & Peak torque for gravitation compensation\\
        First moment of mass & $\SI{5} - \SI{11}{\kilo\gram\meter}$  & To match that of an on-road bicycle \\
        Mass moment of inertia & $\SI{3}- \SI{8}{\kilo\gram\meter\squared}$  & To match that of an on-road bicycle \\
        Closed-loop bandwidth   &  \SI{2}{\hertz} & For accurate torque tracking at target frequencies \\
        Saddle bag height and width & $\SI{0.45}{\meter}$ and $\SI{0.4}{\meter}$ & To hide lean actuation, maintaining the appearance of an on-road bicycle \\
        \hline
    \end{tabular}
\end{table}
\subsection{SPINEA Concept of Operation}
The key concept is a nonlinear geometry that enables both parallel and series action of the springs (see Supplementary Materials Section~\ref{sec:Conceptual Design Steps} for stepwise explanation of the design process). While the springs always connect the bicycle to the motor crank, the misaligned pivot points of the bicycle and of the crank (points $O$ and $M$ in Fig.~\ref{fig:Isometric_View}(a), respectively) yield varying spring force components that work in parallel or in series with the motor, depending on the lean angle of the bicycle and the crank angle. 

Specifically, the left (l) and right (r) spring forces decompose according to their effect (Fig.~\ref{fig:Isometric_View}a): $F_{\mathrm{l}, \bot}$, $F_{\mathrm{r}, \bot}$ are orthogonal to the crank and thus act in series with the motor, as in a conventional SEA. Components $F_{\mathrm{l}, \parallel}$, $F_{\mathrm{r}, \parallel}$ are aligned with the motor crank, so they exert no moment on the crank, but they do exert a moment on the bicycle.  So, a parallel elastic effect arises from the misaligned rotation axes. 

The parallel moment, due to system geometry, is zero  when both bicycle and crank are upright. As lean angles increase, the parallel force components grow, compensating for gravitational effects even if the motor does not exert a moment on the crank. This nonlinear lever effect for the PEA functionality reduces motor torque requirements and acts as a passive safety feature for the rider. 
\subsection{Simulator Components and Realization}
\begin{figure*}
	\centering
	\includegraphics[width = 0.7\textwidth]{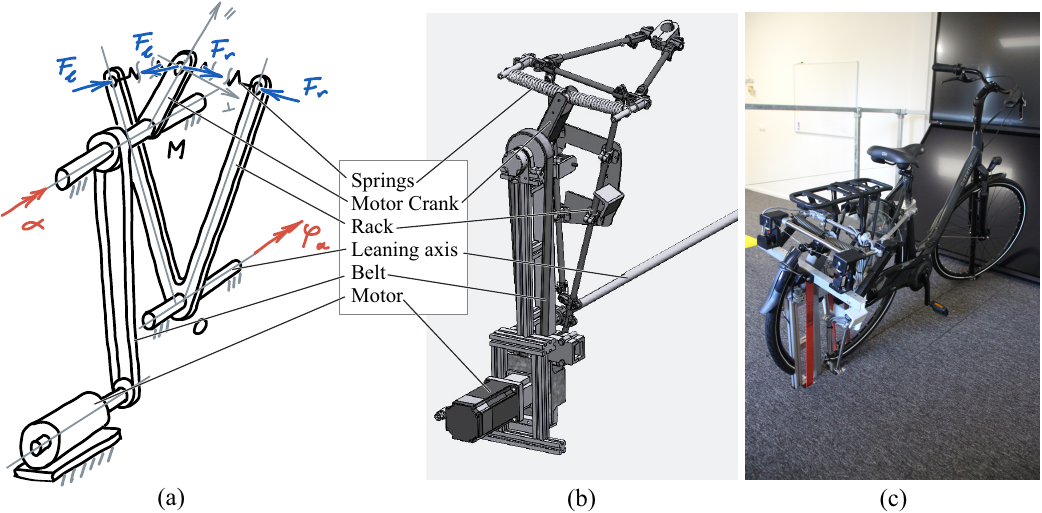}
	\caption{Isometric rear views of SPINEA: (a) Schematic representation of the mechanism, highlighting key components such as the motor, belt drive, crank, series-parallel springs, and bicycle rack. The lean angle of the bicycle rack $\varphi_{a}$ and the crank angle $\alpha$, depicted in red, are measured relative to the upright configuration. Blue arrows represent the left spring force $F_{\mathrm{l}}$ and the right spring force $F_{\mathrm{r}}$, and grey symbols indicate the force decomposition into directions parallel ($\parallel$) and orthogonal ($\bot$) to the crank; (b) Rendering of the final design; (c) Realization of the design on an Orange Bicycle (Gazelle, Dieren, the Netherlands) as part of the bicycle simulator setup at the University of Groningen (photo taken by the author with permission of the host institution). }
	\label{fig:Isometric_View}
\end{figure*}

\begin{figure}
	\centering
    \includegraphics[width=0.75\columnwidth]{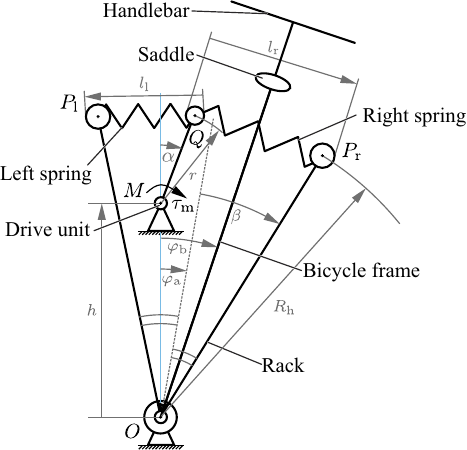}
	\caption{Schematic back view of SPINEA in a leaned position. The motor crank pivots about point $M$ and has a length $r$. It connects to the bicycle rack via two springs of lengths $l_{\mathrm{l}}$ and $l_{\mathrm{r}}$. The bicycle rack is represented by two lines measuring $R_{\mathrm{h}}$ and enclosing an angle of $2\beta$. Each spring hooks onto one pin at the motor crank (point $Q$) and to another pin at points $P_{\mathrm{l}}$ and $P_{\mathrm{r}}$. The bearing location for the bicycle and its rack is point $O$. The centers of rotation $M$ and $O$ are distanced by the height $h$. The grey dashed line denotes the symmetry axis of the bicycle rack. Even though the rack is very stiff and firmly attached to the bicycle frame, this connection is not fully rigid. Instead, the lean angle $\varphi_{\mathrm{a}}$ of the bicycle rack deviates from the lean angle $\varphi_{\mathrm{b}}$ of the bicycle frame measured at the saddle.}
	\label{fig:Actuator_Design_Geometrics}
\end{figure}

In SPINEA's final design (Fig.~\ref{fig:Isometric_View}(a)) the motor crank and the leaning axis, with angles $\alpha$ and $\varphi_{\alpha}$, respectively, are separated by an offset $h$, which is $2.9$ times larger than the motor crank radius $r$ (Fig.~\ref{fig:Actuator_Design_Geometrics}). 
A pair of symmetric antagonistic coil springs T33190 (Tevema, Almere, the Netherlands) facilitate identical system behavior for leaning right and left. Each spring (spring constant $k$) connects the motor crank (attachment point $Q$) to the bicycle frame via a reinforced bicycle ``rack'' (attachment points $P_{\mathrm{l}}$ and $P_{\mathrm{r}}$ in Fig.~\ref{fig:Actuator_Design_Geometrics}). Both rack attachment points ($P_{\mathrm{l}}$ and $P_{\mathrm{r}}$) are located at a distance $R_\mathrm{h}$ from $O$ and placed at an angle~$\beta$ relative to the bicycle’s centerline (Fig.~\ref{fig:Actuator_Design_Geometrics}). 

To actuate the lean, a servo motor drives a compact two-stage transmission including a planetary gearbox and a timing belt. The resulting transmission of 1:24 further reduces the required motor torque, and the timing belt allows the motor to fit beneath the platform. The motor is a SMH100 (Parker-Hannifin Corporation, Cleveland, USA) brushless motor driven by a Parker PSD1 servo drive and equipped with an AB090-010 planetary gearbox from the same manufacturer. From here on, we lump the combination of motor, gearbox, and belt stage into one ``drive unit'' powering a ``motor crank'', with nominal torque $\tau_\mathrm{nom} = \SI{139.2}{\newton\meter}$ and reflected inertia $J_\mathrm{du} = \SI{0.227}{\kilo\gram\meter\squared}$, both calculated at crank level (see calculation in the Supplementary Materials in Section~\ref{sec:Reflected_inertia_and_nominal_torque_at_crank_level}).

The bicycle simulator operates within TwinCAT 3 (version 3.1.4024.32), which integrates real-time control and data logging. Communication with all sensors and actuators is handled via EtherCAT. All geometric parameters are provided in the Supplementary Materials in Table~\ref{tab:Constant_parameters_supplementary}. These values allowed for most mechanical and electronic components to be hidden underneath the platform or inside the saddle bags, remaining concealed throughout the full RoM.
\section{Sensing and Control of SPINEA}
\label{sec:Control}
\subsection{Sensors}
Five sensors monitor the state of the lean actuation of the bicycle (Fig.~\ref{fig:Actuator_Design_Geometrics}): An absolute motor encoder measures the motor angle and thereby indirectly the crank angle $\alpha$. Two wire potentiometers, positioned parallel to the springs, measure the left and right spring elongations, $\Delta l_\mathrm{l}$ and $\Delta l_\mathrm{r}$, respectively. Two additional absolute encoders redundantly measure $\alpha$ at crank level and the lean angle $\varphi_{\mathrm{b}}$ of the bicycle axis, which is located directly beneath the platform. These redundant sensors are used for supervision but not for control, as they are less collocated with the motor.

\subsection{Sensor Processing}
\label{sec:General_mechanics}
Given spring properties, the resultant torque ${\tau}_{\mathrm{sc}}$ that the springs exert on the crank with respect to point $M$ in counter-clockwise direction can be written as a nonlinear function of the geometry, which is fully defined by the (partially redundant) sensor measurements $\Delta l_{\mathrm{l}}$, $\Delta l_{\mathrm{r}}$, and $\alpha$ (Fig.~\ref{fig:Actuator_Design_Geometrics}):  %
\begin{equation}
    \label{eq:tau_sc}
    {\tau}_{\mathrm{sc}} = {\tau}_{\mathrm{sc}}(\Delta l_{\mathrm{l}}, \Delta l_{\mathrm{r}}, \alpha).
\end{equation}
The full expression is given in (\ref{eq:tau_sc_final}) (Supplementary Materials). When assuming a massless and frictionless drive unit and crank, ${\tau}_{\mathrm{sc}}={\tau}_{\mathrm{m}}$ holds. 

The same sensor information also provides the lean torque ${\tau}_{\mathrm{a}}$ defined as the net torque about point~$O$ that the spring forces $F_{\mathrm{l}}$ and $F_{\mathrm{r}}$ exert on the bicycle:
\begin{equation}
    \label{eq:tau_a}
    {\tau}_{\mathrm{a}} = {\tau}_{\mathrm{a}}(\Delta l_{\mathrm{l}}, \Delta l_{ \mathrm{r}}, \alpha).
\end{equation}
We provide the full analytical description in (\ref*{eq:tau_a_final}) of the Supplementary Materials.

Theoretically, the (also measured) absolute lean angle $\varphi_{\mathrm{b}}$ of the bicycle frame could provide redundant information on the current kinematic configuration. However, in both above equations, we choose not to use this angle, because the bicycle frame deforms and the rack angle $\varphi_{\mathrm{a}}$ does not equal $\varphi_{\mathrm{b}}$. In fact, we account for this deformation by defining the lean angle $\varphi_{\mathrm{a}}$ of the rack as illustrated in Fig.~\ref{fig:Actuator_Design_Geometrics}. There is no dedicated sensor for $\varphi_{\mathrm{a}}$, but we estimate it from potentiometer measurements and the motor encoder. Although it is feasible to compute $\varphi_{\mathrm{a}}$ using only one elongation (either $\Delta l_{\mathrm{l}}$ or $\Delta l_{\mathrm{r}}$) we chose to utilize both measurements to enhance accuracy, leading to the following expression: 
\begin{equation}
        \label{eq_lean_angle}
    	{\varphi}_{\mathrm{a}} = {\varphi}_{\mathrm{a}}(\Delta l_{\mathrm{l}}, \Delta l_{\mathrm{r}}, \alpha). 
\end{equation}
A full and explicit definition of the function is available in (\ref{eq:phi_a_final}) in the Supplementary Materials.
\subsection{Torque Control of SPINEA}
\label{sec:software}
\begin{figure*}
	\centering
	\includegraphics{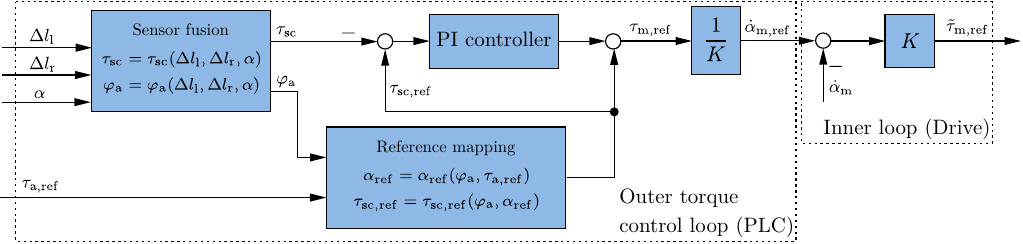}
	\caption{SPINEA control scheme: An outer torque control loop runs on the PLC and commands a reference velocity $\dot{\alpha}_{\mathrm{m, ref}}$ to a fast inner velocity control loop for the motor, running on the motor drive. The reference lean torque on the bicycle rack $\tau_{\mathrm{a, \mathrm{ref}}}$ is an external input, e.g.\ generated by a high-level controller for bicycle dynamics. The PLC algorithm includes: (i) sensor fusion to provide lean angle at rack $\varphi_{\mathrm{a}}$ and crank torque $\tau_{\mathrm{sc}}$,  relying on the left $\Delta l_{\mathrm{l}}$ and right $\Delta l_{\mathrm{r}}$ spring elongations, and the crank angle $\alpha$. (ii) a mapping from reference rack torque to reference crank torque $\tau_{\mathrm{sc, ref}}$, assuming static equilibrium, (iii) the PI controller and (iv) the inverse damping $\frac{1}{K}$, where $K$ can be interpreted as a motor damping coefficient. It transforms the reference motor torque $\tau_{\mathrm{m, ref}}$ into a reference motor velocity $\dot{\alpha}_{\mathrm{m, ref}}$. The drive controls the motor velocity $\dot{\alpha}_{\mathrm{m}}$ by means of a motor reference torque $\tilde{\tau}_{\mathrm{m, ref}}$.}
	\label{fig:Controller}
\end{figure*}

Following conventional force control design for SEAs~\cite{Vallery.2021} that ensures good tracking and low impedance, we implement a cascaded controller (Fig.~\ref{fig:Controller}) that realizes proportional-integral torque control with feedforward in an outer loop, and that leverages the fast velocity control of the motor drive in an inner loop. 
The outer loop runs on a PLC at a sampling rate of $\SI{1}{\kilo\hertz}$, and the inner loop exploits the drive's rapid response with a sampling rate of $\SI{8}{\kilo\hertz}$, similar to \cite{Sensinger.2006, Vallery.2007}. 

The structure of the outer loop resembles that of other SEAs, incorporating both feedforward and feedback components in the force control loop \cite{Pratt.1995}. However, SPINEA's nonlinear kinematics require choosing between closing the loop either in task space (with $\tau_{\mathrm{a}}$) or in crank space (with $\tau_{\mathrm{sc}}$). We choose the latter, to keep nonlinear effects out of the torque feedback loop and be able to rely on simple linear PI control. 

To implement torque control in crank space, we need to map reference torques ${\tau}_{\mathrm{a, ref}}$ for the bicycle lean to reference torques ${\tau}_{\mathrm{sc, ref}}$ for the springs to exert on the crank. That means we need the ``inverse kinematics'' of the nonlinear mechanism, see Fig.~\ref{fig:Controller}. Note that this is not a traditional case of inverse kinematics, because the mechanism is compliant, so $\alpha$ is not known from ${\varphi}_{\mathrm{a}}$. To be able to solve the equations, we assume static equilibrium and massless crank and springs, making the crank reference angle $\alpha_{\mathrm{ref}}$ a static function of ${\varphi}_{\mathrm{a}}$ and ${\tau}_{\mathrm{a, \mathrm{ref}}}$. As we found no closed-form solution, we solve for $\alpha_{\mathrm{ref}}$ online, using a deterministic, iterative bisection interpolation method performed over 20 iterations. This algorithm iteratively adjusts $\alpha_{\mathrm{ref}}$ using ${\varphi}_{\mathrm{a}}$ until the torque $\tau_{\mathrm{a, \mathrm{ref}}}(\varphi_{\mathrm{a}}, \alpha_{\mathrm{ref}})$ converges to match the reference lean torque $\tau_{\mathrm{a, \mathrm{ref}}}$.
Finally, the reference torque $\tau_{\mathrm{sc, ref}}$ results as a function of $\varphi_{\mathrm{a}}$ and $ \alpha_{\mathrm{ref}}$. Intermediate functions are defined in the Supplementary Materials (Section~\ref{sec:Lean_Angle_processing},~\ref{sec:torque_sensing_and_inverse_kinematics} and~\ref{sec:Inverse_kinematics}).

The feedforward component $\tau_{\mathrm{sc,ref}}$ immediately reacts to changes in the reference torque $\tau_{\mathrm{a,ref}}$, enhancing the responsiveness of the system. The proportional-integral (PI) feedback component acts on the error between $\tau_{\mathrm{sc,ref}}$ and $\tau_{\mathrm{sc}}$ via the gains $P_{\tau}$ and $I_{\tau}$.

 As we set the drive's velocity control integral gain to zero, the scheme is mathematically equivalent to PI torque control with added damping with constant $K$ acting on the measured motor velocity $\dot{\alpha}_{\mathrm{m}}$~\cite{Vallery.2021}, i.e.\ the resulting control law  for the reference motor torque $\tilde{\tau}_{\mathrm{m,ref}}$ is: 
\begin{equation}
\label{eq_tau_ref_m}
    \tilde{\tau}_{\mathrm{m,ref}} = \tau_{\mathrm{sc, ref}} + P_{\tau} (\tau_{\mathrm{sc, ref}} - \tau_{\mathrm{sc}}) \\
    + I_{\tau} \int (\tau_{\mathrm{sc, ref}} - \tau_{\mathrm{sc}}) \mathrm{d}t-K\dot{\alpha}_{\mathrm{m}}, 
\end{equation}
with the controller gain chosen as $P_{\tau} = \SI{8}{}$, $I_{\tau} = \SI{50}{\second^{-1}}$ and a motor damping gain
$K =  \SI{27.47}{\newton\meter\second\per\radian}$.
We choose $K$ to be moderate, as we do not aim for velocity control.
\section{Evaluation Methodology}
\label{sec:evaluation methodology}
\subsection{Experimental Setup and Protocol}
We evaluated SPINEA in terms of its transparency (via its torque sensing accuracy, torque tracking performance and impedance characteristics) and output torque range. We conducted three experiments: two that characterized the system without riders, and one with riders cycling on the bicycle simulator. We also conducted supporting simulations.

In the two experiments without riders (\textit{Experiment 1} and \textit{Experiment 2}), we used an external excitation setup and fixed the steering of the bicycle simulator at an angle of $\delta = \SI{0}{\degree}$ (aligning the front wheel with the bicycle frame's symmetry plane). We also attached an S-type load cell DEE, \SI{100}{\kilogram} capacity (Keli, Ostrów Wielkopolski, Poland) to the bicycle frame near the saddle using a rod, to measure the external force $F_{\mathrm{e}}$ acting on the frame. 

\begin{figure*}[hb]
    \centering
	\includegraphics[width=0.8\textwidth]{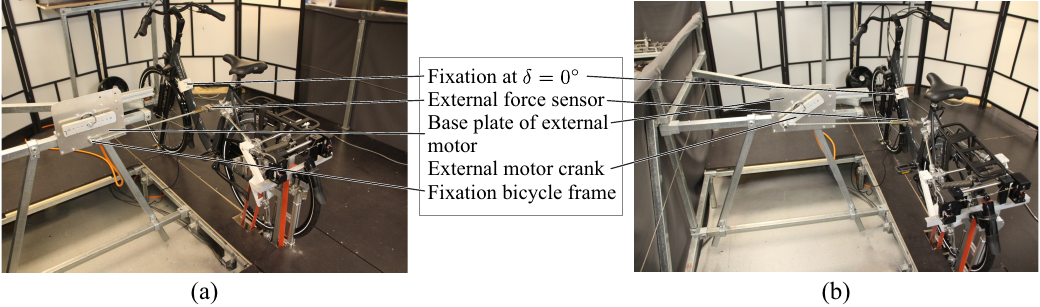}
    \caption{Experimental setups to evaluate torque sensing, torque tracking and disturbance rejection. The steering of the bicycle simulator was locked at an angle of $\delta = \SI{0}{\degree}$ (steering straight). (a) To assess torque sensing and torque tracking, we fixed the bike frame to a rigid base plate while letting the SPINEA track a torque reference $\tau_{\mathrm{a,ref}}$. (b) During the disturbance rejection experiment we connected the bicycle frame via a load cell to the crank of an external motor, which excited the bicycle by tracking a position reference. }
    \label{fig:experiments_external}
\end{figure*}

In \textit{Experiment 1}, we assessed SPINEA's torque sensing and torque tracking performance. We rigidly anchored the sensorized rod to an inertially fixed plate and excited the actuation mechanism via the bicycle motor crank, using $\tau_{\mathrm{a, ref}}$ as the reference (Fig.~\ref{fig:experiments_external}(a)). This setup prevents bicycle frame movement, and so, any influence of the frame's gravitational and inertial effects on the external load cell readings. It also allowed using a synthetic signal for $\tau_{\mathrm{a, \mathrm{ref}}}$, avoiding dependence on dynamic inputs from a rider. 
We chose $\tau_{\mathrm{a, \mathrm{ref}}}$ to be a concatenation of offset-free sinusoidal signals with a maximum amplitude of $\SI{100}{\newton\meter}$. Frequency increased in discrete logarithmic steps from $\SI{0.1}{\Hz}$ to $\SI{7}{\Hz}$, with a total of $18$ frequencies. This protocol ensured the testing of frequencies of interest (up to \SI{2}{\Hz} as per Table~\ref{Tab:Requirements}), while also exploring higher frequencies to further evaluate the actuator's capabilities. The displacement amplitude $A$ decreased with increasing frequency $f$:
\begin{equation}
    A(f) = \min \Big( A_{\mathrm{max}}, \frac{a_{\mathrm{scale}}}{f}A_{\mathrm{max}}\Big),
    \label{eq:Amplitude_Scaling}
\end{equation}
where $ A_{\mathrm{max}}$ is the maximum amplitude and $a_{\mathrm{scale}}$ is the amplitude scaling factor (see Table~\ref{tab:Constant_parameters_excitation}). This scaling limits high-frequency motion while approximately maintaining a constant peak angular velocity ($A(f)2\pi f$). For each amplitude-frequency combination, we collected ten cycles to ensure that the system reached steady-state behavior.

In \textit{Experiment 2}, we assessed the system's impedance characteristics, evaluating its ability to reject external disturbances while tracking zero torque ($\tau_{\mathrm{a,ref}} = \SI{0}{\newton\meter}$). We attached the sensorized rod to an external motor (Fig.~\ref{fig:experiments_external}(b)) and used the external motor’s position as the reference to apply controlled external excitation. This external motor-driven actuation ensures consistent input conditions compared to manual perturbations. 

As the reference for the motor position, we concatenated offset-free sinusoidal signals with increasing frequency and a maximum amplitude of \SI{1}{\radian} (Table~\ref{tab:Constant_parameters_excitation}). Similar to \textit{Experiment~1}, we logarithmically increased the frequency from $\SI{0.1}{\Hz}$ to $\SI{7}{\Hz}$ (using  $18$ intermediate frequencies), applying the same frequency-dependent amplitude scaling as in (\ref{eq:Amplitude_Scaling}).

\textit{Experiment 3} tested riders cycling in the simulator (as in Fig.~\ref{fig:Overview_Simulator}(b)) to evaluate SPINEA's torque capabilities and further assess its torque tracking performance. Data were collected from five adults as part of a larger study approved by the Human Research Ethics Committee at TU Delft. Participants had a mean age of $28.4$ years ($\mathrm{SD} = 3.44$ years) and mean weight of $\SI{76.4}{\kilo\gram}$ ($\mathrm{SD} = \SI{10.86}{\kilo\gram}$). They rode the simulator for at least ten minutes, during which data were recorded on mounting and dismounting, accelerating and braking, and cycling straight both at comfortable and at low speeds.

\subsection{Performance Metrics for Torque Sensing}
To evaluate torque sensing, we used data from \textit{Experiment~1}. To exclude dynamic effects, we used only the low-frequency data of \SI{0.1}{\hertz} by segmenting the first ten cycles, and compared the torque $\tau_{\mathrm{a}}$ at the rack, calculated from \eqref{eq:tau_a}, to the ground-truth torque $\tau_{\mathrm{e}}$ of the external load cell, computed as the product of $F_{\mathrm{e}}$ and the attachment height $h_{\mathrm{e}}$ of the connecting rod. To assess the deviation between $\tau_{\mathrm{a}}$ and $\tau_{\mathrm{e}}$, we report the mean and the root mean square error (RMSE).
\subsection{Identification of Impedance Characteristics}
Using \textit{Experiment 2}, we assessed two types of mechanical impedance: the actuator's and the combined bicycle-actuator system. Actuator impedance should be as small as possible, to make the actuator effectively imperceptible to the user.

We defined the actuator's impedance in frequency domain as $Z_{\mathrm{a}}(j\omega)\coloneqq\tau_{\mathrm{a}}(j\omega)/(j\omega\,\varphi_{\mathrm{a}}(j\omega))$, with the frequency $\omega$. To obtain a theoretical impedance transfer function, we closed the control loop analytically, linearised the equations of motion and solved for $\tau_{\mathrm{a}}(j\omega)/(j\omega\,\varphi_{\mathrm{a}}(j\omega))$ (see Section~\ref*{sec:Linearized transfer functions} of the Supplementary Materials). This provides an analytical expression for $Z_{\mathrm{a}}(j\omega)$ that we can directly compare with the measured impedance.

The simulator's mechanical impedance (i.e., the combined bicycle-actuator) should be similar to that of a real bicycle. We defined this impedance as $Z_{\mathrm{b}}(j\omega)\coloneqq\tau_{\mathrm{e}}(j\omega)/(j\omega\,{\varphi}_{\mathrm{b}}(j\omega))$. To minimize the influence of frame deformation on impedance quantification, we do not use the simulator's lean axis angle sensor to measure ${\varphi_{\mathrm{b}}}$. Instead, we computed  ${\varphi_{\mathrm{b}}}$ from the measured angle $\varphi_{\mathrm{e}}$ of the external excitation motor, as described in the Supplementary Materials (Section~\ref{sec:Lean_angle_sensing_ext_motor}). This positions the sensing of force and displacement in the external excitation mechanism near the bicycle saddle.
To assess whether the simulator met the specifications for first moment of mass and mass moment of inertia about the leaning axis (Table~\ref{Tab:Requirements}), we fitted a linearized model of bicycle leaning with fixed steering axis, reflected inertia $J_{\mathrm{e}}$, gravitational (negative) stiffness $K_{\mathrm{e}}$, and an effective viscous damping $C_{\mathrm{e}}$ to the measured complex impedance $Z_{\mathrm{a}}(j\omega_i)$:
\begin{equation}
\min_{J_{\mathrm{e}},K_{\mathrm{e}},C_{\mathrm{e}}}
\sum_i
 \frac{\left|
Z_{\mathrm{a}}(j\omega_i)
-
\left[
C_{\mathrm{e}}
+
j\left(
\omega_i J_{\mathrm{e}}
-
\frac{K_{\mathrm{e}}}{\omega_i}
\right)
\right]\right|^2
}{
|Z_{\mathrm{a}}(j\omega_i)|^2
}
 ,
\end{equation}
subject to $J_{\mathrm{e}}\ge0$, $C_{\mathrm{e}}\ge0$, and $K_{\mathrm{e}}\le0$. The normalization with $|Z_{\mathrm{a}}(j\omega_i)|$ gives comparable weight to relative deviations in the complex impedance at each excitation frequency. 
\begin{table}
\centering
    \caption{Experimental parameters}
    \label{tab:Constant_parameters_excitation}
    \begin{tabular}{p{4.9cm} p{3.1cm}}
        \hline
        Parameter & Value \\
        \hline
        Maximum amplitude of $\tau_{\mathrm{a, ref}}$ & $A_{\mathrm{max}}(\tau_{\mathrm{a, ref}}) = \SI{100}{\newton\meter}$ \\
        Maximum amplitude $\varphi_{\mathrm{e}}$ & $A_{\mathrm{max}}(\varphi_{\mathrm{e}}) = \SI{1}{\radian}$\\
        Amplitude scaling factor  & $a_{\mathrm{scale}} = 0.2$\\
        Height of fixation of external excitation rod  & $h_{\mathrm{e}} = \SI{0.69}{\meter}$ \\
        \hline
    \end{tabular}
\end{table}
\subsection{Performance Metrics for Torque Tracking} \label{sec:TTP_methods}
We assessed torque tracking performance in experiments with (\textit{Experiment 3}) and without (\textit{Experiment 1}) riders, and compared it to the expected theoretical tracking performance.

First, we linearized equations (\ref{eq:tau_sc}) and (\ref{eq:tau_a}), and used Fig.~\ref{fig:Controller}'s torque control scheme to derive the transfer function ($\tau_{\mathrm{a}}(s)/ \tau_{\mathrm{a, ref}}(s)$) for the torque tracking response (see Section~\ref{sec:Linearized transfer functions} of the Supplementary Materials). This transfer function provided preliminary insights into expected torque tracking performance, which we compared with the data from \textit{Experiment 1} and \textit{Experiment 3}.

We further explored SPINEA's torque tracking performance in two randomly selected participants cycling in the simulator during \textit{Experiment 3}. We identified extended time segments involving straight cycling, resulting in durations of \SI{310}{\second} for rider 1 and \SI{475}{\second} for rider 2. Here, $\tau_{\mathrm{a, \mathrm{ref}}}$ was determined by rider inputs $v$, $\delta$, $\dot{\delta}$, $\varphi_{\mathrm{a}}$ and  $\dot{\varphi}_{\mathrm{a}}$ using the high-level bicycle dynamics controller in eq.~\ref{eq:controller_straight} (Section~\ref{sec:Design_requirements}). 

Across all experiments and simulations, we used the magnitude ratio $\vert\tau_{\mathrm{a}}(j\omega)\vert/\vert\tau_{\mathrm{a, \mathrm{ref}}}(j\omega)\vert$ in a Bode plot to quantify how closely $\tau_{\mathrm{a, \mathrm{ref}}}$ matches $\tau_{\mathrm{a}}$. Additionally, we examined the phase shift between these signals. 

To analyze the cycling data, we first examined the normalised Power Spectral Density (PSD) to estimate the effective bandwidth of the signal (see Supplementary Materials Section~\ref{sec:PSD}). We then determined the dominant frequency of $\tau_{\mathrm{a, \mathrm{ref}}}$ by applying the Fast Fourier Transform (FFT) with windowing techniques (3-second windows) to the time-series data. Next, we calculated the phase shifts and magnitude ratios of $\tau_{\mathrm{a, \mathrm{ref}}}$ and $\tau_{\mathrm{a}}$. Each time window was detrended by subtracting the mean value and multiplied with a Hann window. This approach divided the data into intervals suitable for frequency analysis, minimizing spectral leakage. We report median values of the resulting phase shifts and magnitude ratios for the observed frequency range, to reduce potential measurement outliers. 
\subsection{Performance Metrics for the Torque Range} 
We first explored SPINEA's torque capabilities across the target lean range of motion (RoM) (Table~\ref{Tab:Requirements}) via simulation. For a grid of rack lean angles ${\varphi}_{\mathrm{a}}\in[\SI{-20}{\degree},\SI{20}{\degree}]$ and torques ${\tau}_{\mathrm{a}}\in[\SI{-500}{\newton\meter},\SI{500}{\newton\meter}]$, we calculated the required motor torque $\tau_{\mathrm{m}}$ and crank angle $\alpha$. 

Subsequently, using the logged data from all five riders in \textit{Experiment 3}, we extracted $\varphi_{\mathrm{a}}$ and $\tau_{\mathrm{a}}$ from the measured signals and paired them sample-wise with the simultaneously recorded motor torque $\tau_{\mathrm{m}}$. To facilitate comparison with the simulated motor torque map, we binned the data into discrete classes of $\tau_{\mathrm{m,k}} \in \{0,\pm10,\pm20,\pm30,\pm40,\pm50\}\,\SI{}{\newton\meter}$. For each class, we selected samples satisfying 
$|\tau_{\mathrm{m}}(t)-\tau_{\mathrm{m,k}}| < \SI{1}{\newton\meter}$. We overlaid the resulting ($\varphi_{\mathrm{a}}$, $\tau_{\mathrm{a}}$) samples onto the simulated torque contours. 

We report the peak values for ($\varphi_{\mathrm{a}}$, $\tau_{\mathrm{a}}$) and the measured $\tau_{\mathrm{m}}$ relative to the nominal torque limit $\tau_{\mathrm{nom}}$, and give a qualitative comparison of the measured torque with the simulated contours. This is to (i) evaluate SPINEA's torque capabilities during a cycling task and (ii) validate the simulation results within the RoM covered in the experiments. 
\section{Evaluation Results}
\label{sec:results}
\subsection{Torque Sensing}
\begin{figure}[h]
    \centering
	\includegraphics{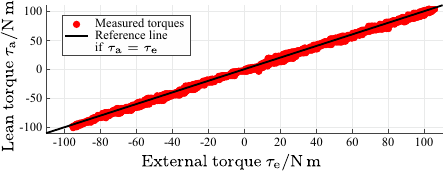}
	\caption{Comparison of the measured lean torque at bicycle rack, $\tau_{\mathrm{a}}$, plotted against the measured external load cell torque, $\tau_{\mathrm{e}}$. The experiments involved applying a sinusoidal reference lean torque at a maximal amplitude of $\tau_{\mathrm{a,ref}} = \SI{100}{\newton\meter}$. The \SI{45}{\degree} identity line (black line) indicates perfect agreement.
    }
	\label{fig:sensor_validation}
\end{figure}

Low-frequency data from \textit{Experiment 1} showed that the RMSE between $\tau_{\mathrm{e}}$ and $\tau_{\mathrm{a}}$ is \SI{3.4}{\newton\meter}. This discrepancy includes a bias component, as the mean values of the two signals differ, with \SI{2.4}{\newton\meter} for $\tau_{\mathrm{e}}$ and \SI{-6e-3}{\newton\meter} for $\tau_{\mathrm{a}}$. This can be seen in Fig.~\ref{fig:sensor_validation}, where $\tau_{\mathrm{a}}$ measurements tend to lie slightly below $\tau_{\mathrm{e}}=\tau_{\mathrm{a}}$.
\subsection{Impedance Characteristics}
\begin{figure}
  \centering
  \includegraphics{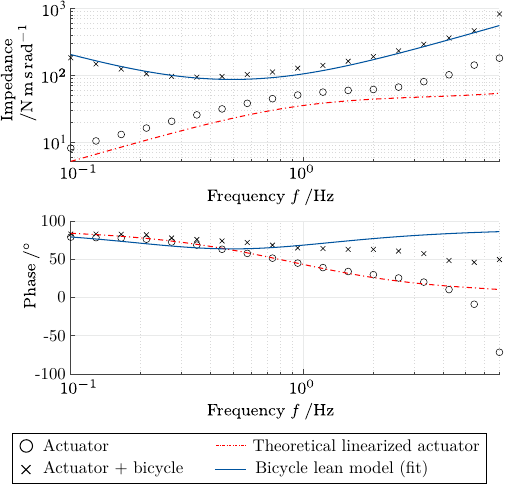}
  \caption{Magnitude and phase of the measured actuator impedance, the combined bicycle‑actuator impedance, the theoretical impedance obtained from the linearised model, and the impedance of a bicycle lean model fit (with inertial, damping, and gravitational elements). 
  }
  \label{fig: Impedance Result}
\end{figure}

\textit{Experiment 2} results show low measured actuator impedance at low frequencies (Fig.~\ref{fig: Impedance Result}). For example, at a frequency of \SI{0.1}{\hertz} the measured actuator impedance is \SI{8.2}{\newton\meter\second\per\radian}. In contrast, the impedance of the combined system is higher (\SI{183}{\newton\meter\second\per\radian}), while the theoretically predicted impedance ((\ref{eq:Z_imp_theory}) in the Supplementary Materials) is lower (\SI{5.2}{\newton\meter\second\per\radian}).

As expected for a SEA and from the theoretical transfer function, the measured actuator impedance follows an inertial trend for low frequencies~\cite{Vallery.2008}, showing an approximate $+\SI{1}{\newton\meter\second\squared\per\radian}$ magnitude slope and a phase angle close to $\SI{90}{\degree}$. However, the measured impedance remains consistently higher than the theoretical prediction.

In the frequency range of $\SI{0.1}{\Hz}$ until $\SI{5.45}{\Hz}$, the measured actuator impedance decreases to a phase angle of $-\SI{8.9}{\degree}$ at $\SI{5.45}{\Hz}$. At $f=\SI{7}{\hertz}$, the phase drops to $-\SI{71.6}{\degree}$. At this excitation frequency, the amplitude of $\varphi_{\mathrm{a}}$ is $\SI{1.4e-4}{\radian}$, and the time signal of $\varphi_{\mathrm{a}}$ no longer shows a clearly discernible sinusoidal oscillation. Accordingly, the $\SI{7}{\hertz}$ frequency-response point was excluded from further analysis.

The magnitude plot of the combined bicycle-actuator system shows a slope of around $-\SI{1}{\newton\meter\second^2\per\radian}$ and a phase angle of $\SI{82.8}{\degree}$ at low frequencies (Fig.~\ref{fig: Impedance Result}), indicating negative stiffness due to gravity. At approximately \SI{0.45}{\Hz}, the slope changes to $+\SI{1}{\newton\meter\second^2\per\radian}$, suggesting that inertial effects dominate. 

For the impedance fit, we excluded the \SI{7}{\hertz} point because $\varphi_{\mathrm{b}}$ did not show a stationary sinusoidal oscillation. The peak amplitudes varied from cycle to cycle, making the extracted amplitude and phase dependent on the selected time window. Parameter identification results using the linearized bicycle leaning model are a lean inertia $J_{\mathrm{e}} = \SI{12.5}{\kilo \gram \,\meter^2}$, a first moment of mass $mx_\mathrm{c} = K_{\mathrm{e}} / g = \SI{12.3}{\kilo \gram \meter}$, and a damping coefficient $C_{\mathrm{e}} = \SI{38.8}{\newton\meter\second\per\radian}$.

\subsection{Torque Tracking Performance} \label{sec:TTP_results}
\begin{figure*}
	\centering
	\includegraphics[width = 0.95\textwidth]{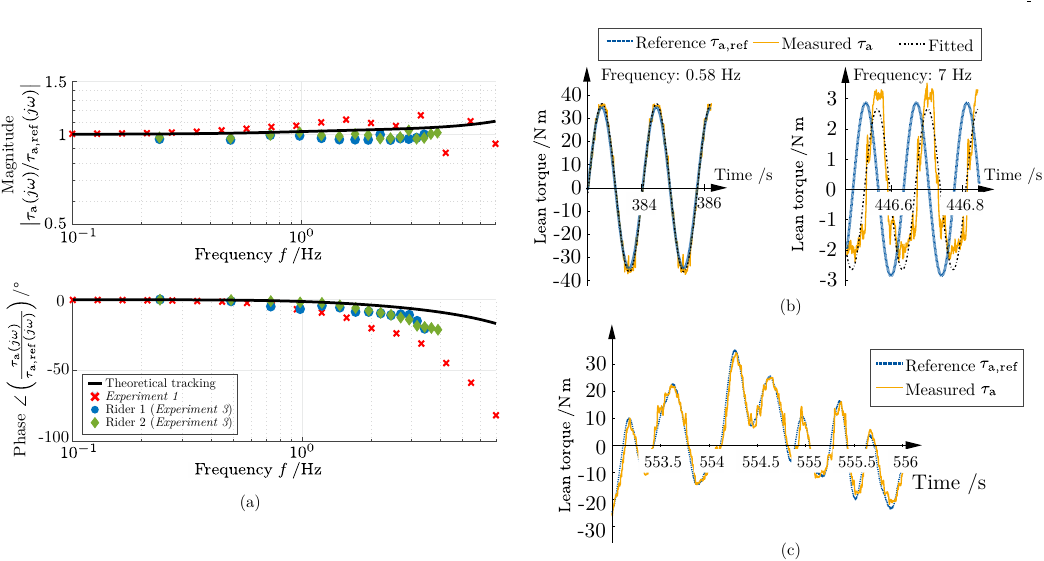}
	\caption{Results of \textit{Experiments 1} and \textit{3}. (a) Bode plot with the magnitude and phase shift of the measured lean torque at the bicycle rack, $\tau_{\mathrm{a}}$, in relation to the reference lean torque,  $\tau_{\mathrm{a, \mathrm{ref}}}$. The results from \textit{Experiment 1}, during which the bicycle frame was fixed, are indicated with red crosses. Riders 1 and 2 from \textit{Experiment 3} are represented by filled blue circle and green diamond markers, respectively. The solid black line depicts the theoretical tracking performance as per the linearized model of SPINEA. (b) A time series snippet from \textit{Experiment 1} is presented, illustrating, $\tau_{\mathrm{a, \mathrm{ref}}}$ and $\tau_{\mathrm{a}}$ for two different excitation frequencies. (c) A time series snippet for \textit{Experiment 3}, Rider 1, is shown, depicting both $\tau_{\mathrm{a, \mathrm{ref}}}$ and $\tau_{\mathrm{a}}$.}
  \label{fig:Force_Tracking_Real}
\end{figure*}

In \textit{Experiments 1} and \textit{3}, the magnitude does not decrease by \SI{3}{\decibel} below the maximum measured magnitude, as typically used to define bandwidth (Fig.~\ref{fig:Force_Tracking_Real}(a)). \textit{Experiment 1} results show a minimum magnitude of \SI{0.87} and a maximum magnitude of \SI{1.16}, while in \textit{Experiment 3} they range between \SI{0.96} and \SI{1.03}. Therefore, we defined a conservative control bandwidth as the frequency at which the phase shift is \SI{45}{\degree}.

In \textit{Experiment 1}, a phase shift of \SI{45}{\degree} occurs at \SI{4.25}{\Hz}. Fig.~\ref{fig:Force_Tracking_Real}(b) shows example results for a reference of \SI{0.58}{\Hz}, showing only minimal tracking errors. For frequencies higher than \SI{4.25}{\Hz}, \textit{Experiment 1} results indicate decreased tracking performance with increasing phase shifts (Fig.~\ref{fig:Force_Tracking_Real}(a)). At the upper frequency limit of our reference signal, \SI{7}{\Hz}, the phase shift reaches $-\SI{81.2}{\degree}$. This decreased performance is also evident in the time domain (Fig.~\ref{fig:Force_Tracking_Real}(b)). In contrast, the theoretically derived transfer function (Fig.~\ref{fig:Force_Tracking_Real}(a)) displays consistent magnitude alignment with experiments but shows less pronounced changes in phase shift at higher frequencies, culminating in a phase shift of only $-\SI{16.8}{\degree}$ at \SI{7}{\Hz}.

The phase shift in \textit{Experiment 3} never reached \SI{45}{\degree}, suggesting good torque tracking performance across all recorded frequencies. The cycling task encompassed frequencies of up to \SI{4}{\hertz}. Fig.~\ref{fig:Force_Tracking_Real}(c) shows a snippet of Rider 1 time series data. Interestingly, Fig.~\ref{fig:Force_Tracking_Real}(a) suggests that although phase shift trends are similar across experiments (increasing with frequency), human rider experiments (\textit{Experiment 3}) show better torque tracking performance, as indicated by lower phase shifts compared to \textit{Experiment 1}. 
\subsection{Torque Capabilities}
Open-loop simulations for a \SI{100}{\kilo\gram} rider including the \SI{21}{\kilo\gram} bicycle show a maximum gravitational bike torque $\tau_{\mathrm{g,max}}$ of around \SI{365}{\newton\meter}, requiring only about \SI{10}{\newton\meter} from the motor at crank level. This $\tau_{\mathrm{g,max}}$ (Fig.~\ref{fig:Real_Torque_Profile}) occurs at the edges of the workspace (lean angle of $\pm \SI{20}{\degree}$), while the required $\tau_{\mathrm{m}}$ remains within the nominal torque capacity $\tau_{\mathrm{nom}}$ due to the parallel effect of the springs.

\textit{Experiment 3} results also show increasing $\tau_{\mathrm{a}}$ for larger $\varphi_{\mathrm{a}}$, but reach values that would exceed $\tau_{\mathrm{nom}}$. For example, at $\varphi_{\mathrm{a}}=-\SI{8.35}{\degree}$, the bike torque is $\tau_{\mathrm{a}} = \SI{157.9}{\newton\meter}$. However, SPINEA attains these levels with motor torques $\tau_{\mathrm{m}}$ well below $\tau_{\mathrm{nom}}$.

Compared to simulation outcomes, higher torques $\tau_{\mathrm{a}}$ are achieved at smaller $\tau_{\mathrm{m}}$ values with riders cycling in \textit{Experiment 3}. This is evident from instances where $\tau_{\mathrm{m}}$ lies above contour lines of corresponding $\tau_{\mathrm{sc}}$. This phenomenon is most noticeable for torque values of $\tau_{\mathrm{m}} = \pm \SI{30}{\newton\meter}$ when compared with $\tau_{\mathrm{sc}} = \pm \SI{30}{\newton\meter}$ (Fig.~\ref{fig:Real_Torque_Profile}).

\section{Discussion}
\label{sec:Discussion}

\subsection{SPINEA Shows Accurate Torque Tracking}
We observed accurate torque tracking up to frequencies of \SI{4.25}{\hertz} during \textit{Experiment 1}, and up to \SI{4}{\hertz} (the maximum observed frequency) during \textit{Experiment 3}, meeting our target of achieving a control bandwidth of $\SI{2}{\hertz}$ (Table~\ref{Tab:Requirements}).

Interestingly, reduced phase shifts were recorded during \textit{Experiment 3}, with riders cycling, compared to \textit{Experiment~1}. We initially attributed the improved phase response to changing dynamics arising from nonzero lean angle of the bicycle, but simulations indicated that this effect would be smaller. An alternative explanation could be that human riders act as phase-correcting elements within the control loop. Here, when providing accurate steering inputs in response to lean dynamics, they effectively enhance system responsiveness.

\subsection{Torque Sensing is Precise, but Sensitive to Spring Elongation Measurement}
Our torque sensing evaluation, comparing lean torque at the bicycle rack with external torque measurements, showed an RMSE of \SI{3.4}{\newton\meter}. A likely source of the observed error is measurement inaccuracy in $\tau_{\mathrm{a}}$. The potentiometers have a sensing precision of \SI{0.25}{mm} according to their data sheet. Due to the high spring stiffness, \SI{0.25}{mm} error in elongation measurement translates to lean‑torque errors at the bicycle rack around \SI{1.5}{\newton\meter}. 

\subsection{Simulator Resembles a High-Inertia Bike with Added Damping}
The impedance analysis revealed that at low frequencies, the actuator exhibited negligible inertial characteristics compared to the bicycle, such that the combined system behaved like a ``negative-stiffness'' element (Fig.~\ref{fig: Impedance Result}). This suggests that with zero reference lean torque, impedance experienced by the rider realistically arises from gravitational effects on the bicycle. These results, along with the torque tracking results, support SPINEA's transparency at low frequencies.

The apparent actuator impedance exceeds the theoretically expected impedance. This difference may stem from static friction in the drive train. Also, above \SI{3.3}{\hertz}, magnitude and phase indicate that stiffness effects become more prominent. 

To assess whether SPINEA led to realistic lean dynamics, we modeled the combined bicycle-actuator system as a leaning bicycle with fixed steering axis and gravitational, inertial and damping components. The experimental identification of the inertia of our system's lean DoF yielded $J_{\mathrm{e}} = \SI{12.5}{\kilo\gram\meter\squared}$, which we compared to values reported in bicycle literature \cite{Kooijman.2008,Kalsbeek_2016}, ranging from $J = \SI{3} - \SI{8}{\kilo\gram\meter\squared}$ (Table~\ref{Tab:Requirements}).

\begin{figure}[ht]
	\centering
    \includegraphics{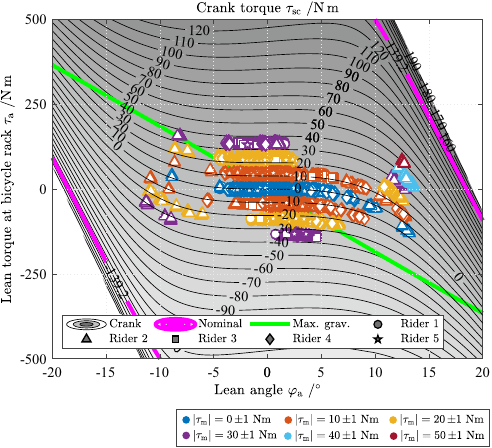}
	\caption{Crank torque $\tau_{\mathrm{sc}}$ (grey contour lines) as a function of lean angle $\varphi_{\mathrm{a}}$ and lean torque $\tau_{\mathrm{a}}$. Pink lines indicate the motor's nominal torque $\tau_{\mathrm{nom}}$ and the green line the maximum needed lean torque $\tau_{\mathrm{g,max}}$ to compensate gravity for a rider having a mass of \SI{100}{\kilo\gram} and the bicycle of \SI{21}{\kilogram}. Rider data from \textit{Experiment 3} is depicted with colored markers for different discrete $\tau_{\mathrm{m}}$. Here, data points are shown only when $\tau_{\mathrm{m}}$ is within $\pm \SI{1}{\newton\meter}$ of discrete torque levels in steps of $\SI{10}{\newton\meter}$. Each rider has a different marker shape.}
	\label{fig:Real_Torque_Profile}
\end{figure}

Besides the actuator, also additional components that lean with the bicycle contribute to inertia, particularly the propulsion motor and gearbox. Following literature on inertia for other bicycle geometries, a similarly sized electric bicycle would be expected to reach an inertia in the range of $J = \SI{8} - \SI{11}{\kilo\gram\meter\squared}$\cite{Kalsbeek_2016}, still slightly below our identified value. This places the inertia of our system's lean DoF between that of a riderless electric bicycle and that of a riderless motorcycle ($J \approx \SI{45.7}{\kilo\gram\meter\squared}$ \cite{Passigato.2020}), still below that of a city bicycle with a rider ($J \approx \SI{78.05}{\kilo\gram\meter\squared}$ \cite{Meijaard.2007}). 

In terms of the first moment of mass (mass times centre of mass height) around the lean axis, our simulator yields $\SI{12.3}{\kilo\gram\meter}$. This falls within the range reported for comparable electric bicycles (i.e., $\SI{12} - \SI{18}{\kilo\gram\meter}$ \cite{Kalsbeek_2016}). The discrepancy can be attributed primarily to differences in mass distribution.

While on-road bicycle models include no explicit lean damping (i.e., $\M{C_1}(1,1) = 0$ \cite{Meijaard.2007}), our identification indicates a non-zero \emph{effective} damping of \SI{38.8}{\newton\meter\second\per\radian} for simulator lean. We attribute this mainly to friction in the over-constrained lean axis assembly (with seven bearings, Fig.~\ref{fig:Isometric_View}). 

It should be noted that the  impedance analysis assumes that SPINEA operates sufficiently linearly within the tested frequency range, particularly at small lean angles. We confirm this assumption in our Supplementary Materials (Fig.~\ref{fig:Impedance_Offset}), finding consistent phase and magnitude responses when the system is excited both off-centre and in the upright configuration. 

In terms of appearance and usability, SPINEA facilitated the leaning motion of the simulator on a stable platform while remaining hidden throughout its full range of motion. This actuator design allowed the bicycle simulator to closely resemble the appearance of an on-road bicycle.

\subsection{SPINEA Delivers the Needed Torques across the RoM}
SPINEA was able to deliver the high lean torque outputs required for its application in the bicycle simulator (Table~\ref{Tab:Requirements}), counteracting gravitational forces without overloading the motor at the edges of the workspace ($\pm\SI{20}{\degree}$) during simulations, and up to \SI{13.6}{\degree} with riders cycling on the bicycle simulator (Fig.~\ref{fig:Real_Torque_Profile}). Note that riders never leaned more than \SI{13.6}{\degree}, likely because they cycled straight and did not fall. For the same motor torque input, the lean torque at the bicycle was larger in human rider experiments than predicted by the simulation, suggesting modeling inaccuracies like neglected friction in the transmission.

\subsection{Future Work and Other Uses of SPINEA}
Results showed acceptable torque tracking and actuator impedance, so 
we now intend to evaluate the simulator's ability to replicate real-world bicycle dynamics via on-road comparison studies. Here, assessing the system in various training and rehabilitation scenarios could further establish its applicability and effectiveness as a tool for enhancing cycling skills and rehabilitation outcomes.

Additionally, future work could include more precise sensing of spring elongation, to further improve the transparency and realism of the system. 

The SPINEA concept also holds promise for other applications requiring actuators with both high torques and high torque-tracking fidelity. We anticipate the biggest gains in mobile or wearable robotics, where minimal weight of the overall system is more critical than in a stationary system. For example, the key design feature of the misaligned axes may be a helpful modification to support SEA robotic arms against gravity without requiring additional parallel elastic elements.

For some applications, SPINEA can be further simplified, for example by letting go of the symmetry requirement. If symmetry is not needed, a single spring could suffice. 

\section{Conclusion}
\label{sec:conclusion}
This work introduced SPINEA, a series-parallel integrated nonlinear elastic actuator designed to meet the conflicting requirements of delivering high torque at specific RoM areas, while maintaining transparency and safety.

Our evaluation results indicated that SPINEA effectively incorporates features of SEAs and PEAs through its nonlinear geometry. It maintains transparency at low frequencies, showing good tracking performance (up to \SI{4.25}{\hertz}) and low impedance, while being capable of delivering high torques at the workspace edges and ensuring safety through passive gravitational compensation. 

While this work applied SPINEA to a bicycle simulator, it also provides insights for broader applications in actuator designs that benefit from combined series and parallel effects. 

\section*{Acknowledgments}
We thank Jan van Frankenhuyzen for his work on the mechanics and experimental setup; Dick de Waard, Bastiaan Sporrel, Arjan Stuiver, and Riender Happee for their support with the actuation requirements, human study, and simulator photographs; Philipp Schubert for feedback on earlier drafts; and Royal Dutch Gazelle for providing the bicycle.

\bibliographystyle{IEEEtran}
\bibliography{bibliography}

@inproceedings{Pratt.1995,
 author = {Pratt, G. A. and Williamson, M. M.},
 title = {Series elastic actuators},
 pages = {399--406},
 publisher = {{IEEE Comput. Soc. Press}},
 isbn = {0-8186-7108-4},
 booktitle = {Proceedings 1995 IEEE/RSJ International Conference on Intelligent Robots and Systems. Human Robot Interaction and Cooperative Robots},
 year = {1995},
 doi = {10.1109/IROS.1995.525827}
}

@inproceedings{Sensinger.2006,
 author = {Sensinger, Jonathon and {Ff. Weir}, Richard},
 title = {Improvements to Series Elastic Actuators},
 pages = {1--7},
 publisher = {IEEE},
 isbn = {0-7803-9721-5},
 booktitle = {2006 2nd IEEE/ASME International Conference on Mechatronics and Embedded Systems and Applications},
 year = {2006},
 doi = {10.1109/MESA.2006.296927}
}

@inproceedings{Vallery.2007,
 author = {Vallery, Heike and Ekkelenkamp, Ralf and {van der Kooij}, Herman and Buss, Martin},
 title = {Passive and accurate torque control of series elastic actuators},
 pages = {3534--3538},
 publisher = {IEEE},
 isbn = {978-1-4244-0911-2},
 booktitle = {2007 IEEE/RSJ International Conference on Intelligent Robots and Systems},
 year = {2007},
 doi = {10.1109/IROS.2007.4399172}
}

@article{Wintersberger.2022,
 author = {Wintersberger, Philipp and Matviienko, Andrii and Schweidler, Andreas and Michahelles, Florian},
 year = {2022},
 title = {Development and Evaluation of a Motion-based VR Bicycle Simulator},
 pages = {1--19},
 volume = {6},
 number = {MHCI},
 journal = {Proceedings of the ACM on Human-Computer Interaction},
 doi = {10.1145/3546745}
}

@inproceedings{MartinezGarcia.2023,
 author = {{Martinez Garcia}, Donaji and Gr{\"o}ne, Kilian and Fischer, Martin and Zhao, Min and Bergen, Melina},
 title = {Validation of a bicycle simulator based on objective criteria},
 publisher = {{The Evolving Scholar - BMD 2023, 5th Edition}},
 booktitle = {The Evolving Scholar - BMD 2023, 5th Edition},
 year = {2023},
 doi = {10.59490/650d765b1dbdfef75c1c3a17}
}

@article{Meijaard.2007,
 author = {Meijaard, J.P and Papadopoulos, Jim M. and Ruina, Andy and Schwab, A.L},
 year = {2007},
 title = {Linearized dynamics equations for the balance and steer of a bicycle: a benchmark and review},
 pages = {1955--1982},
 volume = {463},
 number = {2084},
 issn = {1364-5021},
 journal = {Proceedings of the Royal Society A: Mathematical, Physical and Engineering Sciences},
 doi = {10.1098/rspa.2007.1857}
}

@inproceedings{Plooij.2015,
 author = {Plooij, Michiel and {van Nunspeet}, Marvin and Wisse, Martijn and Vallery, Heike},
 title = {Design and evaluation of the Bi-directional Clutched Parallel Elastic Actuator (BIC-PEA)},
 pages = {1002--1009},
 publisher = {IEEE},
 isbn = {978-1-4799-6923-4},
 booktitle = {2015 IEEE International Conference on Robotics and Automation (ICRA)},
 year = {2015},
 doi = {10.1109/ICRA.2015.7139299}
}

@article{Toxiri.2018,
 author = {Toxiri, Stefano and Calanca, Andrea and Ortiz, Jesus and Fiorini, Paolo and Caldwell, Darwin G.},
 year = {2018},
 title = {A Parallel-Elastic Actuator for a Torque-Controlled Back-Support Exoskeleton},
 pages = {492--499},
 volume = {3},
 number = {1},
 journal = {IEEE Robotics and Automation Letters},
 doi = {10.1109/LRA.2017.2768120}
}

@inproceedings{Grimmer.2012,
 author = {Grimmer, Martin and Eslamy, Mahdy and Gliech, Stefan and Seyfarth, Andre},
 title = {A comparison of parallel- and series elastic elements in an actuator for mimicking human ankle joint in walking and running},
 pages = {2463--2470},
 publisher = {IEEE},
 isbn = {978-1-4673-1405-3},
 booktitle = {2012 IEEE International Conference on Robotics and Automation},
 year = {2012},
 doi = {10.1109/ICRA.2012.6224967}
}

@article{Roozing.2021,
 author = {Roozing, Wesley and Ren, Zeyu and Tsagarakis, Nikos G.},
 year = {2021},
 title = {An efficient leg with series--parallel and biarticular compliant actuation: design optimization, modeling, and control of the eLeg},
 pages = {37--54},
 volume = {40},
 number = {1},
 issn = {0278-3649},
 journal = {The International Journal of Robotics Research},
 doi = {10.1177/0278364919893762}
}

@article{Pfeifer.2015,
 author = {Pfeifer, Serge and Pagel, Anna and Riener, Robert and Vallery, Heike},
 year = {2015},
 title = {Actuator With Angle-Dependent Elasticity for Biomimetic Transfemoral Prostheses},
 pages = {1384--1394},
 volume = {20},
 number = {3},
 issn = {1083-4435},
 journal = {IEEE/ASME Transactions on Mechatronics},
 doi = {10.1109/TMECH.2014.2337514}
}

@article{Shoman.2021,
 author = {Shoman, Murad M. and Imine, Hocine},
 year = {2021},
 title = {Bicycle Simulator Improvement and Validation},
 pages = {55063--55076},
 volume = {9},
 journal = {IEEE Access},
 doi = {10.1109/ACCESS.2021.3071214}
}

@inproceedings{Kwon.2001,
 author = {Kwon, Dong-Soo and Yang, Gi-Hun and Lee, Chong-Won and Shin, Jae-Cheol and Park, Youngjin and Jung, Byungbo and Lee, Doo Yong and Lee, Kyungno and Han, Soon-Hung and Yoo, Byoung-Hyun and Wohn, Kwang-Yun and Ahn, Jung-Hyun},
 title = {KAIST interactive bicycle simulator},
 pages = {2313--2318},
 publisher = {IEEE},
 isbn = {0-7803-6576-3},
 booktitle = {Proceedings 2001 ICRA. IEEE International Conference on Robotics and Automation (Cat. No.01CH37164)},
 year = {2001},
 doi = {10.1109/ROBOT.2001.932967}
}

@incollection{Pratt.1997,
 author = {Pratt, Gill A. and Williamson, Matthew M. and Dillworth, Peter and Pratt, Jerry and Wright, Anne},
 title = {Stiffness isn't everything},
 pages = {253--262},
 volume = {223},
 publisher = {Springer-Verlag},
 isbn = {3-540-76133-0},
 series = {Lecture Notes in Control and Information Sciences},
 editor = {Khatib, Oussama and Salisbury, J. Kenneth},
 booktitle = {Experimental Robotics IV},
 year = {1997},
 address = {London},
 doi = {10.1007/BFb0035216}
}

@incollection{Vallery.2021,
 author = {Vallery, H. and Plooij, M.},
 title = {Compliant Multi-DOF Actuation},
 pages = {45--54},
 publisher = {{Springer International Publishing}},
 isbn = {978-3-030-40886-2},
 editor = {Beckerle, Philipp and Sharbafi, Maziar Ahmad and Verstraten, Tom and Pott, Peter P. and Seyfarth, Andr{\'e}},
 booktitle = {Novel Bioinspired Actuator Designs for Robotics},
 year = {2021},
 address = {Cham},
 doi = {10.1007/978-3-030-40886-2{\textunderscore }5}
}

@article{Vallery.2008,
 author = {Vallery, Heike and Veneman, Jan and {van Asseldonk}, Edwin and Ekkelenkamp, Ralf and Buss, Martin and {van der Kooij}, Herman},
 year = {2008},
 title = {Compliant actuation of rehabilitation robots},
 pages = {60--69},
 volume = {15},
 number = {3},
 issn = {1070-9932},
 journal = {IEEE Robotics {\&} Automation Magazine},
 doi = {10.1109/MRA.2008.927689}
}

@article{Wyss.2019,
 author = {Wyss, Dario and Pennycott, Andrew and Bartenbach, Volker and Riener, Robert and Vallery, Heike},
 year = {2019},
 title = {A MUltidimensional Compliant Decoupled Actuator (MUCDA) for Pelvic Support During Gait},
 pages = {164--174},
 volume = {24},
 number = {1},
 issn = {1083-4435},
 journal = {IEEE/ASME Transactions on Mechatronics},
 doi = {10.1109/TMECH.2018.2878289}
}

@article{Talbourdet.1941,
 author = {{Talbourdet G. J.}},
 year = {1941},
 title = {Mathematical solution of four-bar linkages},
 volume = {13},
 number = {5-7},
 journal = {{Machine Design}}
}

@misc{Kalsbeek_2016,
  author       = {Inge M. Kalsbeek},
  title        = {Experimental investigation into the shimmy motion of the bicycle for improving model-based shimmy estimations},
  year         = {2016},
  howpublished = { http://resolver.tudelft.nl/uuid:a98d51c1-7754-4c29-b883-f130ba05136b},
  note         = {TU Delft Institutional Repository},
}

@article{Passigato.2020,
 author = {Passigato, Francesco and Eisele, Andreas and Wisselmann, Dirk and Gordner, Achim and Diermeyer, Frank},
 year = {2020},
 title = {Analysis of the Phenomena Causing Weave and Wobble in Two-Wheelers},
 pages = {6826},
 volume = {10},
 number = {19},
 journal = {Applied Sciences},
 doi = {10.3390/app10196826 }
}

@article{Wolf.2016,
 author = {Wolf, Sebastian and Grioli, Giorgio and Eiberger, Oliver and Friedl, Werner and Grebenstein, Markus and Hoppner, Hannes and Burdet, Etienne and Caldwell, Darwin G. and Carloni, Raffaella and Catalano, Manuel G. and Lefeber, Dirk and Stramigioli, Stefano and Tsagarakis, Nikos and {van Damme}, Michael and {van Ham}, Ronald and Vanderborght, Bram and Visser, Ludo C. and Bicchi, Antonio and Albu-Schaffer, Alin},
 year = {2016},
 title = {Variable Stiffness Actuators: Review on Design and Components},
 pages = {2418--2430},
 volume = {21},
 number = {5},
 issn = {1083-4435},
 journal = {IEEE/ASME Transactions on Mechatronics},
 doi = {10.1109/TMECH.2015.2501019}
}

@article{Paine.2014,
 author = {Paine, Nicholas and Oh, Sehoon and Sentis, Luis},
 year = {2014},
 title = {Design and Control Considerations for High-Performance Series Elastic Actuators},
 pages = {1080--1091},
 volume = {19},
 number = {3},
 issn = {1083-4435},
 journal = {IEEE/ASME Transactions on Mechatronics},
 doi = {10.1109/TMECH.2013.2270435}
}

@article{Verstraten.2016,
 author = {Verstraten, Tom and Beckerle, Philipp and Furn{\'e}mont, Rapha{\"e}l and Mathijssen, Glenn and Vanderborght, Bram and Lefeber, Dirk},
 year = {2016},
 title = {Series and Parallel Elastic Actuation: Impact of natural dynamics on power and energy consumption},
 pages = {232--246},
 volume = {102},
 issn = {0094114X},
 journal = {Mechanism and Machine Theory},
 doi = {10.1016/j.mechmachtheory.2016.04.004}
}

@article{Krimsky.2024,
 author = {Krimsky, Erez and Collins, Steven H.},
 year = {2024},
 title = {Elastic energy-recycling actuators for efficient robots},
 pages = {eadj7246},
 volume = {9},
 number = {88},
 journal = {Science robotics},
 doi = {10.1126/scirobotics.adj7246}
}

@article{Liu.2024,
 author = {Liu, Siyu and Ding, Jiatao and Lu, Chunlei and Wang, Zhirui and Su, Bo and Guo, Zhao},
 year = {2024},
 title = {A Novel Optimization Design of Dual-Slide Parallel Elastic Actuator for Legged Robots},
 pages = {2886--2894},
 volume = {29},
 number = {4},
 issn = {1083-4435},
 journal = {IEEE/ASME Transactions on Mechatronics},
 doi = {10.1109/TMECH.2024.3401546}
}

@mastersthesis{VandenOuden.2011 ,
  author       = {J.H. Van den Ouden },
  title        = {Inventory of Bicycle Motion for the Design of a Bicycle Simulator},
  school       = {Delft University of Technology},
  year         = {2011},
  type         = {Master's thesis},
  url          = {https://resolver.tudelft.nl/uuid:ec31182d-8063-41a3-89ec-799be901cb6e}
}

@article{Cain.2016,
 author = {Cain, Stephen M. and Ashton-Miller, James A. and Perkins, Noel C.},
 year = {2016},
 title = {On the Skill of Balancing While Riding a Bicycle},
 pages = {e0149340},
 volume = {11},
 number = {2},
 journal = {PloS one},
 doi = {10.1371/journal.pone.0149340}
}

@article{Sanjurjo.2019,
 author = {Sanjurjo, Emilio and Naya, Miguel A. and Cuadrado, Javier and Schwab, Arend L.},
 year = {2019},
 title = {Roll angle estimator based on angular rate measurements for bicycles},
 pages = {1705--1719},
 volume = {57},
 number = {11},
 issn = {0042-3114},
 journal = {Vehicle System Dynamics},
 doi = {10.1080/00423114.2018.1551554}
}

@article{Moore.2011,
 author = {Moore, Jason K. and Kooijman, J. D. G. and Schwab, A. L. and Hubbard, Mont},
 year = {2011},
 title = {Rider motion identification during normal bicycling by means of principal component analysis},
 pages = {225--244},
 volume = {25},
 number = {2},
 issn = {1384-5640},
 journal = {Multibody System Dynamics},
 doi = {10.1007/s11044-010-9225-8}
}

@article{Kooijman.2008,
 author = {Kooijman, J. D. G. and Schwab, A. L. and Meijaard, J. P.},
 year = {2008},
 title = {Experimental validation of a model of an uncontrolled bicycle},
 pages = {115--132},
 volume = {19},
 number = {1-2},
 issn = {1384-5640},
 journal = {Multibody System Dynamics},
 doi = {10.1007/s11044-007-9050-x}
}

@inproceedings{Colgate.1989,
 author = {Colgate, E. and Hogan, N.},
 title = {An analysis of contact instability in terms of passive physical equivalents},
 pages = {404--409},
 publisher = {{IEEE Comput. Soc. Press}},
 isbn = {0-8186-1938-4},
 booktitle = {Proceedings, 1989 International Conference on Robotics and Automation},
 year = {1989},
 doi = {10.1109/ROBOT.1989.100021}
}

@inproceedings{moore2010accurate,
  author    = {Moore, Jason K. and Hubbard, Mont and Schwab, A. L. and Kooijman, J. D. G.},
  title     = {Accurate Measurement of Bicycle Parameters},
  booktitle = {Proceedings of the Bicycle and Motorcycle Dynamics 2010 Symposium on the Dynamics and Control of Single Track Vehicles},
  address   = {Delft, The Netherlands},
  month     = oct,
  year      = {2010},
  pages     = {},
}

@article{Mathijssen.2015,
 author = {Mathijssen, Glenn and Lefeber, Dirk and Vanderborght, Bram},
 year = {2015},
 title = {Variable Recruitment of Parallel Elastic Elements: Series--Parallel Elastic Actuators (SPEA) With Dephased Mutilated Gears},
 pages = {594--602},
 volume = {20},
 number = {2},
 issn = {1083-4435},
 journal = {IEEE/ASME Transactions on Mechatronics},
 doi = {10.1109/TMECH.2014.2307122}
}

\setcounter{figure}{0}
\renewcommand{\thefigure}{S\arabic{figure}}

\setcounter{table}{0}
\renewcommand{\thetable}{S\arabic{table}}

\setcounter{equation}{0}
\renewcommand{\theequation}{S\arabic{equation}}

\setcounter{section}{0}
\renewcommand{\thesection}{S\arabic{section}}

\onecolumn

\section*{Supplementary Materials}

\section{Overview of the Supplementary Materials}
This document provides additional derivations, parameter values, and experimental analyses that complement the main manuscript. First, we detail design steps (Design (a)–(e)) that progressively evolved into our final symmetric SPINEA concept. Next, we compute the reflected inertia and nominal torque at crank level to characterize the drive unit as a single equivalent actuator. We then detail the lean and steer actuation model implemented on the bicycle simulator. Afterwards, we introduce the four-bar baseline mechanism used as a rigid reference for SPINEA and describe the associated optimization procedure. Subsequent sections present the full geometric relations for lean-angle sensing (with an external motor and on the bicycle), the torque sensing formulation, and the steady-state reference mapping between rack torque and crank torque for SPINEA. Finally, we derive linearized transfer functions for impedance and torque tracking and report additional experimental results, including power spectral densities of the cycling task and Bode plot analyses that support the assumption of local linearity.

\section{Conceptual Design Steps}
\label{sec:Conceptual Design Steps}
\begin{figure*}[h]
    \centering
    \includegraphics[width=0.9\linewidth]{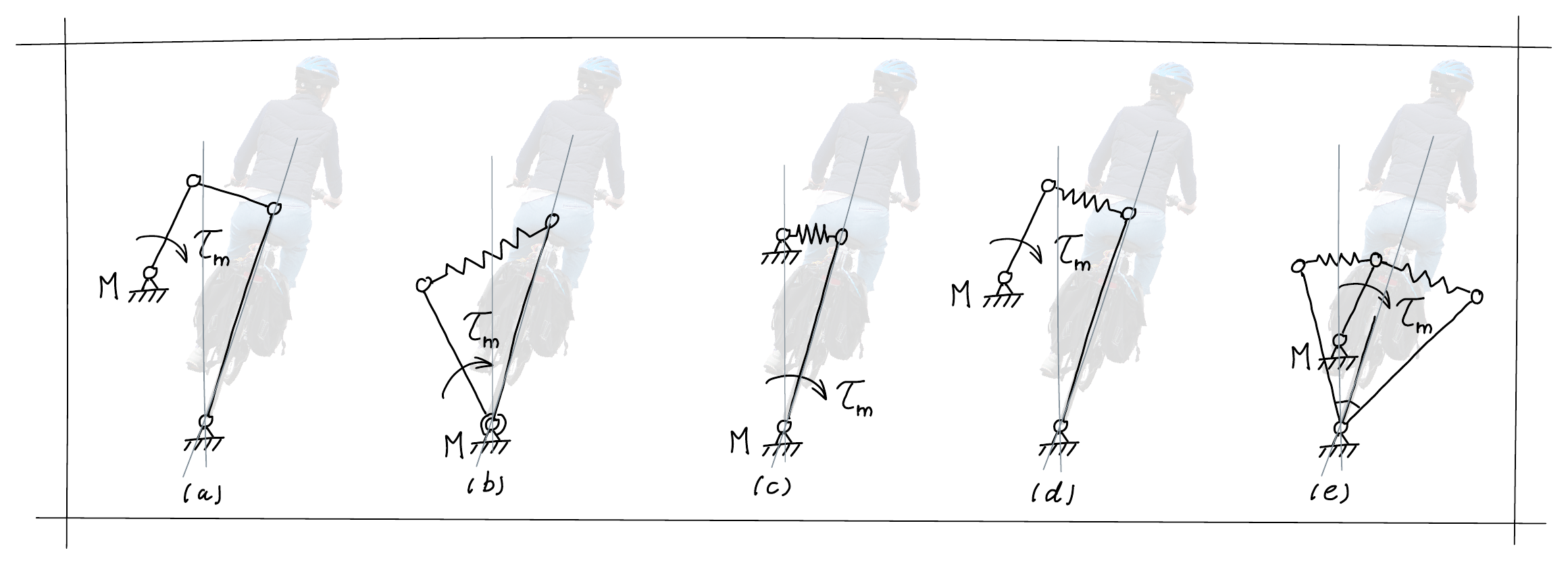}
    \caption{Schematic view of five distinct lean actuation mechanisms in a deflected position. The location of motor actuation is indicated by $M$. (a) Four-bar mechanism featuring an offset motor position, consisting of four linkages. (b) Series elastic actuator: The rotational axis of both the motor and bicycle leaning axis share the same rotational axis passing through point $M$ in the frontal plane. The motor crank is connected to the bicycle via a series spring. (c) Parallel elastic actuator: A spring between the bicycle and a fixed anchor complements the motor torque. (d) Asymmetric SPINEA design: This design substitutes the coupler from the four-bar mechanism with a spring, providing both series and parallel spring effect. (e) Symmetric SPINEA concept}
    \label{fig:Lean_Actuation_Concept}
\end{figure*}

With the requirements in mind, we developed five actuation concepts: Designs (a) through (e) as depicted in Fig.~\ref{fig:Lean_Actuation_Concept}. We started with Design (a), iterating on intermediate steps (b) to (d), and culminating in Design (e) as our final concept.  

Design (a) (Fig.~\ref{fig:Lean_Actuation_Concept}(a)) uses a four-bar mechanism that optimizes bar lengths for defined rotational motion and reduces motor torque requirements through mechanical advantage. It consists of four linkages: the driven linkage (crank) connected to the motor, the following linkage attached to the leaning axis, and a coupler linking both. While it provides mechanical advantage, its rigid connection limits possible reduction of reflected actuator inertia~\cite{Colgate.1989}.

To enhance force control precision and decouple the motor's mechanical impedance from the bicycle, Design (b) (Fig.~\ref{fig:Lean_Actuation_Concept}(b)) follows a conventional SEA concept by integrating a spring in series with motor actuation. It effectively extends the driven linkage to match the follower's length. However, this choice compromises mechanical advantage, requiring the motor and transmission system to supply full output torque, and complicates concealing actuation components in the saddle bags. 

For load sharing, Design (c) connects the motor crank to a fixed anchor via a spring (Fig.~\ref{fig:Lean_Actuation_Concept}(c)). This configuration follows the principle of Parallel Elastic Actuators (PEAs) by allowing the spring to share loads from rider weight, reducing the motor's workload. However, it neither provides mechanical advantage nor directly improves force control. 

Design (d) features a hybrid actuator in which the spring serves dual roles. This is a prototypical SPINEA. By replacing the rigid coupler of the four-bar mechanism with a flexible spring element, this approach allows for nonlinear compliant transmission during bicycle leaning. These nonlinear kinematics provide mechanical advantage tailored to the application and enable the spring to simultaneously complement the motor torque like in a PEA, and to ensure low impedance as in a SEA, all with a minimal number of components. However, this SPINEA setup exhibits asymmetric load distribution across the workspace, which is undesirable for our symmetric setup.

Design (e) balances the SPINEA's load distribution by introducing a symmetric configuration with antagonistic springs (Fig.~\ref{fig:Lean_Actuation_Concept}(e)), facilitating torque control for the symmetric bicycle simulator and mechanical integration into a compact system.

\section{Reflected inertia and nominal torque at crank level}
\label{sec:Reflected_inertia_and_nominal_torque_at_crank_level}
To characterize the drive unit as a single equivalent actuator at the crank, we compute the reflected inertia and the nominal torque at crank level. This allows us to directly verify whether the required crank torque exceeds the nominal torque of the drive unit. We can express the drive unit inertia reflected to the crank as:
\begin{equation}
    J_{\mathrm{du}} = (J_{\mathrm{m}} + J_{\mathrm{g}}) \, i_{\mathrm{m, sc}}^2 + J_{\mathrm{bs}} + J_{\mathrm{c}}.
\end{equation}
Here, $J_{\mathrm{du}}$ denotes the total inertia of the drive unit referred to the crank level; rotating bearing components are neglected since their inertial properties are unknown. The term $J_{\mathrm{m}}$ represents the motor rotor inertia, while $J_{\mathrm{g}}$ is the inertia of the gearbox input shaft. We define the overall transmission $i_{\mathrm{m, sc}}$ from motor to crank as the product of the gearbox ratio $i_{\mathrm{g}}$ and the belt ratio $i_{\mathrm{bs}}$. The remaining parts $J_{\mathrm{bs}}$ and $J_{\mathrm{c}}$ account for the inertia of the belt stage and the inertia of the rotating crank, respectively. 

We estimate the crank inertia and the belt-stage inertia from the pulley inertias obtained from the computer-aided design (CAD) model. The 72-tooth pulley is connected to the gearbox, while the 30-tooth pulley is connected to the crank. We refer both pulley inertias to the crank side and obtain:
\begin{equation}
   J_{\mathrm{bs}} = J_{\mathrm{p72}} + i_{\mathrm{bs}}^2 J_{\mathrm{p30}} + m_{\mathrm{belt}} r_{\mathrm{p72}}^2.
\end{equation}
To obtain the corresponding nominal torque at crank level, the motor torque is mapped through the total transmission ratio \(i_{\mathrm{m,sc}}\). In this work, a lossless drivetrain is assumed, i.e., losses in the gear stage, belt, and bearings are neglected. The nominal torque at crank level is therefore given by
\begin{equation}
    \tau_{\mathrm{nom}} = \tau_{\mathrm{m,nom}} \, i_{\mathrm{m,sc}}  .
\end{equation}
The numerical values used to compute the reflected inertia and the nominal torque at crank level are summarized in Table~\ref{tab:drivetrain_params}.
\begin{table}[h]
\centering
    \caption{Parameters of the drive unit}
    \label{tab:drivetrain_params}
    \begin{tabular}{p{6cm} p{4.5cm}}
        \hline
        Parameter & Value \\
        \hline
        Motor rotor inertia & $J_{\mathrm{m}} = \SI{3.36e-4}{\kilogram\meter\squared}$\\
        Gearbox ratio & $i_{\mathrm{g}} = 10$\\
        Inertia of the gearbox & $J_{\mathrm{g}} = \SI{4.4e-5}{\kilogram\meter\squared}$ \\
        Belt ratio & $i_{\mathrm{bs}} = \frac{72}{30}$\\
        Inertia of 30-tooth pulley & $J_{\mathrm{p30}}= \SI{3.788e-5}{\kilogram\meter\squared}$\\
        Inertia of 72-tooth pulley & $J_{\mathrm{p72}}= \SI{1.675e-3}{\kilogram\meter\squared} $\\
        Radius of 72 tooth pulley & $r_{\mathrm{p72}}= \SI{0.0567}{\meter} $\\
        Mass of belt & $m_{\mathrm{belt}}= \SI{0.144}{\kilogram} $\\
        Inertia of motor crank & $J_{\mathrm{c}} = \SI{5.8e-3}{\kilogram\meter\squared}$ \\
        Nominal ungeared motor torque from datasheet & $\tau_{\mathrm{m,nom}} = \SI{5.8}{\newton\meter}$\\
        \hline
    \end{tabular}
\end{table}

\section{High-level controller to generate the reference torque}
\label{sec:Lean_and_steer_torque_actuation_model}
Here we present the model, and the parameter values used, for the estimated SPINEA's {torque requirements}. For the model we use the linearised equations of motion for the lean and steer angle of the bicycle, as presented by Meijaard et al. \cite{Meijaard.2007},
\begin{equation}
\begin{split}
	\M{M} \begin{pmatrix}
    \ddot{\varphi} \\
    \ddot{\delta}
    \end{pmatrix}+v\M{C}_1 \begin{pmatrix}
    \dot{\varphi} \\
    \dot{\delta}
    \end{pmatrix} + \left(g \M{K}_0 + v^2 \M{K}_2 \right)\begin{pmatrix}
    {\varphi} \\
    {\delta}
    \end{pmatrix}=\begin{pmatrix}
    \tau_\varphi \\
    \tau_\delta
    \end{pmatrix} ,
\end{split}
	\label{eq:suppl_eom_real}
\end{equation}
with the lean angle $\varphi$,  steer angle $\delta$, and the applied lean torque $\tau_\varphi$ and  steer torque $\tau_\delta$. The matrices $\M{M},\M{C}_1, \M{K}_0, \M{K}_2$ are constants derived from the bicycle geometry and mass distribution, $g$ is the gravitational constant and $v$ the forward velocity.  

The bicycle simulator actuated by SPINEA consists of an off-the-shelf Orange C7+ Bicycle (Royal Dutch Gazelle, Dieren, the Netherlands) fixed in the longitudinal direction, but capable of operating lean and steer motions independently (Fig.~\ref{fig:Overview_Simulator}(a) of the main manuscript). We assume that the simulator shares the same mass and gravitational properties as the bicycle on the street and that the lean and steer torque actuators only need to render the velocity dependent terms from the equations of motion. However, the bicycle simulator setup has zero trail at the front wheel, because the front assembly support is located on the extended steering axis at ground level. This means that there is no steer to yaw coupling and that the static or gravitational steer to lean coupling is only partly present. Therefore, we have added some off-diagonal gravitational stiffness to the lean and steer actuation, based on the linearized equations of motion. The reference leaning torque $\tau_{\mathrm{a, \mathrm{ref}}}$ and the reference steering torque $\tau_{\delta, \mathrm{ref}}$ can, after correction for the trail, be expressed as,
\begin{equation}
\begin{split}
\begin{pmatrix}
    \tau_{\mathrm{a, \mathrm{ref}}} \\
    \tau_{\delta, \mathrm{ref}}
\end{pmatrix}
&= -\Bigg(
    v \M{C}_1 
    \begin{pmatrix}
        \dot{\varphi} \\
        \dot{\delta}
    \end{pmatrix} 
 +
    \left( v^2 \M{K}_2 + g 
    \begin{pmatrix}
        0 & K_{0, \mathrm{a}, \varphi \delta} \\ 
        K_{0, \mathrm{a}, \delta \varphi} & 0
    \end{pmatrix} 
    \right) 
    \begin{pmatrix}
        {\varphi} \\ 
        {\delta}
    \end{pmatrix}
\Bigg).
\end{split}
\label{eq:suppl_controller_straight}
\end{equation}
We calculated $\M{C}_1, \M{K}_2$ and the off-diagonal terms in $\M{K}_0$ for our Gazelle bicycle using the equations from Meijaard et al.~\cite{Meijaard.2007} and the geometric and mass parameters reported for "Bicycle C" by Kalsbeek \cite{Kalsbeek_2016} (see Table~~\ref{Tab:Lean_and_steer_actuation_model_parameters}). Bicycle C is an earlier Gazelle model with comparable characteristics to those of our Orange 7+ model. Initially, we used for the additional off-diagonal gravitational stiffness terms $K_{0, \mathrm{a},\delta\varphi} = K_{0, \delta\varphi}$ and $K_{0, \mathrm{a},\varphi \delta} = K_{0, \varphi\delta}$. However, during pilot testing, the simulator was felt to be too aggressive in the lean-to-steer coupling, especially at higher forward speeds. To limit this behaviour, we capped the forward velocity $v$ in (\ref{eq:suppl_controller_straight}) to \SI{4}{\meter\per\second} and reduced $K_{0, \mathrm{a},\delta\varphi}$ by 50\% (see Table~\ref{Tab:Lean_and_steer_actuation_model_parameters}). 

\begin{table}[h]
\centering
    \caption{Lean and steer actuation model parameters}
    \label{Tab:Lean_and_steer_actuation_model_parameters}
    \begin{tabular}{p{4cm} p{5cm}}
        \hline
        Component & Value \\
        \hline
        Gyroscopic matrix $\M{C_1}$ & $\M{C_1} = \begin{pmatrix}
            0 & 30.5822 \\
            -0.4823 & 1.4912
        \end{pmatrix}$ \si{\kilogram\meter}\\
        Centrifugal matrix $\M{K_2}$ & $\M{K_2} = \begin{pmatrix}
            0 & 71.9171 \\
            0 & 2.1023
        \end{pmatrix}$ \si{\kilogram\second} \\
       Gravitational off-diagonal term & $K_{0, \varphi \delta} = -2.3570$ \si{\kilogram\meter}\\
        Gravitational off-diagonal term & $K_{0, \delta \varphi} = -1.1785$ \si{\kilogram\meter}\\
     \end{tabular}
\end{table}

\section{Comparison Four-Bar Rigid Actuator and SPINEA}
\label{sec:Comparison_Four_Bar}
\begin{figure*}[ht]
	\centering
	\includegraphics{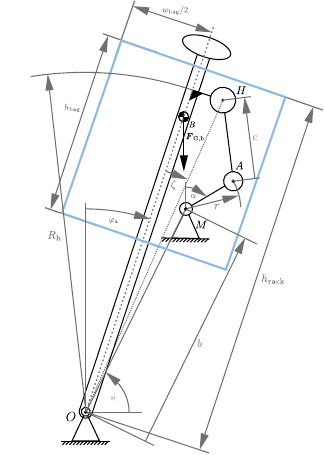}
	\caption{Four-bar mechanism with 6 design parameters: radius $R_{\mathrm{h}}$ between point $H$ and point $O$, length $b$ of the fixed link, coupler length $c$, radius $r$ of the driven linkage (crank), angle $\upsilon$ between the base linkage and the horizontal axis, and offset angle $\zeta$ between the bicycle's symmetry axis and a line from $O$ to the connection point $H$ of the coupler). Geometric constraints are formulated such that connection points $A$ (connecting driven linkage with coupler) and $M$ (connecting driven linkage to motor) stay within the bag (defined by the width $w_{\mathrm{bag}}$ and height $h_{\mathrm{bag}}$) and below the rack height $h_{\mathrm{rack}}$ throughout the range of motion. The crank angle $\alpha$ varies, causing the bicycle rack angle $\varphi_{\mathrm{a}}$ to change within the range of $- \SI{20}{\degree}$ and $+ \SI{20}{\degree}$. The gravitational force $F_{\mathrm{G, b}}$ acts in the center of mass $B$ of the bicycle frame. The parameters resulting from the optimization are listed in Table~\ref{tab:Constant_parameters_supplementary}.}
  \label{fig:Four_bar}
\end{figure*}

To provide a baseline for comparison with the proposed SPINEA concept, we use a standard four-bar mechanism (Design~(a)), whose design incorporates several critical considerations to ensure effective transmission. A key aspect is maintaining nearly 90-degree angles between the lever arm and the coupler, which enhances the performance of the mechanism. The lengths of all links - denoted as $R_{\mathrm{h}}$, $b$, $c$, $r$ - along with the angle $\upsilon$ and the offset angle $\zeta$ are freely selectable parameters (Fig.~\ref{fig:Four_bar}). Here, $R_{\mathrm{h}}$ represents the radius between point $H$ and point $O$, $b$ is the length of the fixed link, so the distance between the bearings of the bicycle rack and motor crank, $c$ is length of coupler between point $A$ and $H$ and $r$ the radius of the driven linkage (crank). The angle $\upsilon$ denotes the orientation between the line connecting point $O$ and $H$ to the horizontal axis, while $\zeta$ indicates the offset angle between the bicycle's symmetry axis and the line connecting $O$ and $H$.

We first defined a feasible baseline (pre-optimization) four-bar geometry using Talbourdet’s kinematic relations from 1941, and geometric constraints ensuring that the linkage connection points remain within the saddlebag volume and below the rack height for the entire lean range of motion \cite{Talbourdet.1941}. This baseline served as the initial guess for a constrained nonlinear optimization (MATLAB fmincon, SQP), which tuned the freely selectable parameters to minimize the maximum required motor torque over the RoM of the lean angle.

As in Design (a), Design (e) includes a motor that actuates a crank connected to the bicycle. However, Design (e) integrates compliant elements, resulting in a two-degree-of-freedom system. 

For both designs (a) and (e), we considered all components (links and springs) massless except for the output link --- the bicycle. For Design (e), we derived the equations of motion (Section~\ref{sec:torque_sensing_and_inverse_kinematics}).

\begin{figure*}
	\centering
	\includegraphics{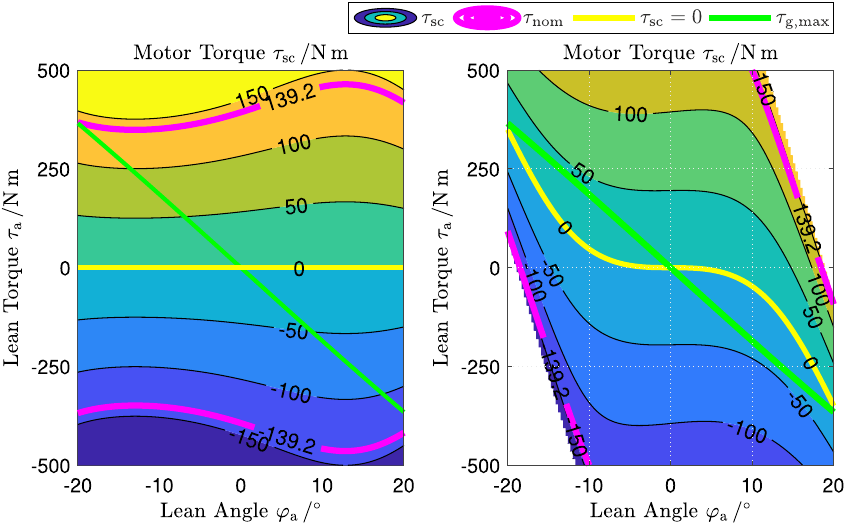}
	\caption{Comparison of the two mechanical designs using contour plots to illustrate torque performance for the full range of motion of $\pm \SI{20}{\degree}$ ($\tau_{\mathrm{sc}}$: Crank torque, $\tau_{\mathrm{nom}}$: Nominal torque of the drive unit, $\tau_{\mathrm{sc}}=0$: Zero torque of the drive unit, $\tau_{\mathrm{g, max}}$: Maximal gravitational torque) ; (a) Four-bar mechanism; (b) SPINEA}
  \label{fig:Comparison}
\end{figure*}

A comparison between designs (a) and (e) can be seen in Fig.~\ref{fig:Comparison}. The contour plot illustrates the relationship between lean torque $\tau_{\mathrm{a}}$ measured at bicycle rack and the crank torque $\tau_{\mathrm{sc}}$. We highlight the nominal torque of the drive unit $\tau_{\mathrm{nom}}$. The line $\tau_{\mathrm{g,max}}$ shows the required lean torque for a rider-bicycle system with a mass of \SI{121}{\kilo\gram}. 

Our analysis reveals that both mechanisms maintain the maximum gravitational torque $\tau_{\mathrm{g,max}}$ within the limits of $\tau_{\mathrm{nom}}$. For the four-bar mechanism, the maximal gravitational torque remains below this threshold but approaches it closely at a lean angle of -\SI{20}{\degree}. However, three key differences exist between both designs. 

First, contour lines for Design (a) lack symmetry during bicycle motion relative to its upright position; conversely, Design (e) exhibits inverted symmetry, facilitating improved control during operation. 

Second, when the drive unit provides zero torque ($\tau_\mathrm{m}=0$) in Design (e), lean torque $\tau_\mathrm{a}$ on the bicycle does not automatically equal zero; rather, increasing lean angles correspond with rising lean torques. This effect becomes particularly pronounced at high lean angles---such as 20 degrees---wherein $\tau_{\mathrm{a}}>\tau_{\mathrm{g, max}}$. This suggested that our system could support a rider safely, preventing falls despite isolated leaning motions being inherently unstable on traditional bicycles. In contrast, in Design (a), when the torque of the drive unit is zero, the lean torque is also zero. This presents a potential risk where users could fall if power to the system is interrupted.

Third, the SPINEA's nonlinear design ``shapes'' output torque characteristics to closely match the expected main task requirements (compensating gravity). That results in substantial margin between required and nominal motor torque especially around workspace edges, which could either be exploited for higher dynamic range or to further downsize the motor.

The analysis presented thus far could be extended. For example, additional linear transmission stages could further increase the motor's mechanical advantage in either design. However, increased transmission ratios tend to result in higher friction, greater reflected inertia, and elevated motor speed requirements, which collectively lead to increased power demands.
\section{Lean angle sensing with external motor}
\label{sec:Lean_angle_sensing_ext_motor}
\begin{figure}[ht]
	\centering
	\includegraphics{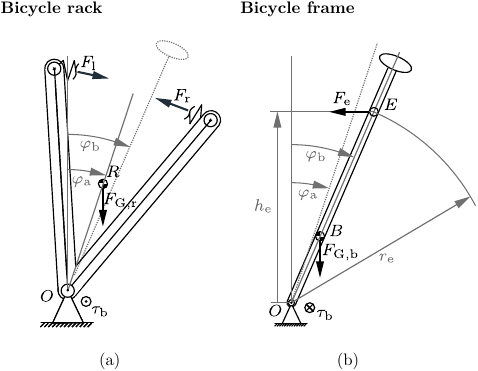}
	\caption{Simplified models of the bicycle rack and frame, with their connection lumped into a bending torque ${\tau}_{\mathrm{b}}$, such that frame deformation can be included in the model. (a) Forces ${F}_{\mathrm{l}}$ and ${F}_{\mathrm{r}}$ acting on the bicycle rack. Deformation causes the rack angle  ${\varphi}_{\mathrm{a}}$  to deviate from the bicycle frame angle $\varphi_{\mathrm{b}}$. The force ${F}_{\mathrm{G,r}}$ indicates the rack’s gravitational force. (b) Model of the bicycle frame, showing the excitation force ${F}_{\mathrm{e}}$ acting at the height $h_{\mathrm{e}}$ and the frame’s gravitational force ${F}_{\mathrm{G, b}}$. In this experiment, the bicycle lean angle $\varphi_{\mathrm{b}}$ is measured by the external excitation motor’s angle.} 
	\label{fig:Angles_impedance}
\end{figure}

In \textit{Experiment 2}, we excited the bicycle with an external motor, which allows to determine the lean angle $\varphi_{\mathrm{b}}$ (see Fig. \ref{fig:Angles_impedance}) of the bicycle frame from the external motor angle $\varphi_{\mathrm{e}}$ via simple kinematic mapping. Specifically, we obtain
\begin{equation}
    {\varphi}_{\mathrm{b}}(t) = \mathrm{asin} \left(\sin({\varphi_{\mathrm{e}}}(t))  \frac{r_{\mathrm{e}}}{h_{\mathrm{e}}}\right) \quad,  
    \label{eq:external_impedance}
\end{equation}
where $r_{\mathrm{e}}$ represents the length of the motor crank of the excitation motor and $h_{\mathrm{e}}$ the attachment height of the connecting rod (Table~\ref{tab:Constant_parameters_supplementary}). 
\section{Lean angle sensing on bicycle}
\label{sec:Lean_Angle_processing}
\begin{figure}
\centering
\includegraphics[width=0.6\textwidth]{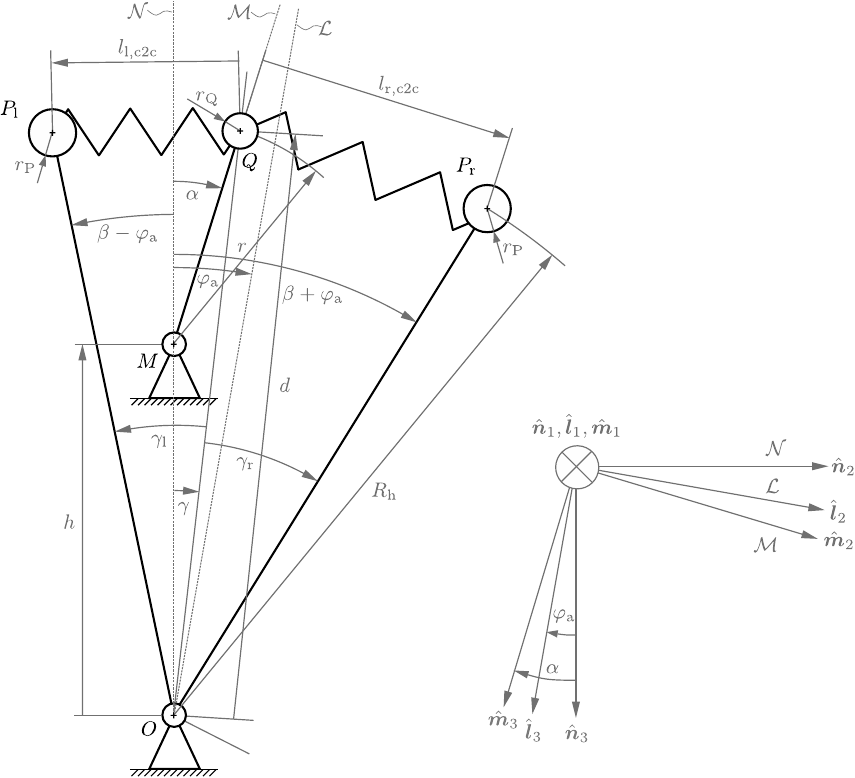}
\caption{Schematic of the lean actuation. In this 2D representation, the motor crank pivots about $M$ and has a length $r$. Two springs hook onto a pin of radius $r_{\mathrm{Q}}$ at Point $Q$ on the crank and respectively connect it to pins of radius $r_{\mathrm{P}}$ at points $P_{\mathrm{l}}$ and $P_{\mathrm{r}}$ on the rack. The rack is represented by two lines measuring $R_{\mathrm{h}}$ and enclosing an angle of $2\beta$. Spring lengths $l_{l,c2c}$ and $l_{r,c2c}$ are measured from center to center of respective pins. The bicycle rack pivots about point $O$, and the grey dashed line denotes its symmetry axis. The two points $M$ and $O$ are distanced by the height $h$. Two triads, $\mathcal{L}$ and $\mathcal{M}$, indicate the two degrees of freedom (DoFs): Triad $\mathcal{L}$ represents the rotational degree of freedom for the bicycle rack measured relative to $\mathcal{N}$ with the lean angle bicycle rack $\varphi_{\mathrm{a}}$, while triad $\mathcal{M}$ corresponds to the rotation of the motor crank measured relative to $\mathcal{N}$ with the crank angle $\alpha$. The angle $\gamma_{\mathrm{l}}$ of the rack's left ``arm'' and the angle $\gamma_{\mathrm{r}}$ of the right ``arm'' with respect to the line $\overline{OQ}$, along with the angle $\gamma$ formed by the distance $d$ between points $O$ and $Q$ and the line $\overline{OM}$, are intermediate values that help make the calculations of $\varphi_{\mathrm{a}}$ more comprehensible. The constant parameters are defined in Table~\ref{tab:Constant_parameters_supplementary}.}
\label{fig:Lean_sensing}
\end{figure}

Although we equipped the lean axis with an angle encoder, in this work we reconstruct the rack lean angle $\varphi_{\mathrm{a}}$ from kinematics using the crank angle measured with the absolute motor encoder and the spring elongations. In particular, we calculate $\varphi_{\mathrm{a}}$ (see Fig.~\ref{fig:Lean_sensing}) from the crank angle $\alpha$ (which the motor control unit measures) and the left and right spring elongations, $\Delta l_{\mathrm{l}}$ and $\Delta l_{\mathrm{r}}$, respectively (which wire potentiometers measure in parallel to the springs).

\begin{figure}[h]
    \centering
    \hfill
    \subfloat{%
        \includegraphics[width=0.32\linewidth]{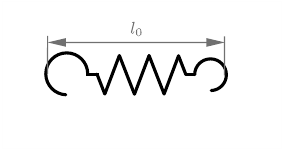}%
        \label{fig:Rest_length}
    }
    \subfloat{%
        \includegraphics[width=0.32\linewidth]{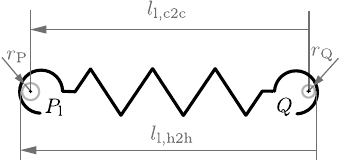}%
        \label{fig:Left_spring_length}
    }
    \subfloat{%
        \includegraphics[width=0.32\linewidth]{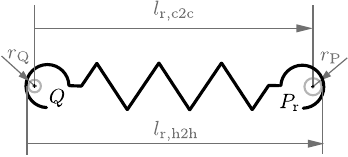}%
        \label{fig:Right_spring_length}
    }
    \caption{Length relationships of springs: rest lengths and hook distances. (a) Rest length $l_0$ of both springs; (b) Left spring length between hook to hook $l_{\mathrm{l,h}2\mathrm{h}}$ is a summation of the left length between the center to center $l_{\mathrm{l,c}2\mathrm{c}}$ and two radii at point $Q$ and point $P_{\mathrm{r}}$ or $P_{\mathrm{l}}$, $r_{\mathrm{Q}}$ and $r_{\mathrm{P}}$; (c) Right spring length between hook to hook $l_{\mathrm{r,h}2\mathrm{h}}$ is a summation of the right length between the center to center $l_{\mathrm{r,c}2\mathrm{c}}$, $r_{\mathrm{Q}}$ and $r_{\mathrm{P}}$.}
    \label{fig:spring_lengths}
\end{figure}

Starting with $\alpha$, we determine the distance $d$ between the origin $O$ and the spring attachment point $Q$ on the crank as:
\begin{equation}
	\label{eq_d}
	d = \sqrt{r^2+h^2+2   r   h   \cos(\alpha) },
\end{equation}
where $h$ the offset of the motor crank axis above the lean axis and $r$ as the radius of the motor crank. From this we can determine the angle $\gamma$ of the line $\overline{OQ}$ with the vertical as:
\begin{equation}
	\label{eq_gamma}
	\gamma = \arcsin   \left(\frac{r}{d}\, \,\sin(\pi-\alpha) \right) .
\end{equation}
The angle $\gamma_{\mathrm{l}}$ of the left arm and the angle $\gamma_{\mathrm{r}}$ of the right arm with respect to the line $\overline{OQ}$ can now be calculated from the left and right spring lengths, $l_{\mathrm{l,c}2\mathrm{c}}$ and $l_{\mathrm{r,c}2\mathrm{c}}$, determined from axle center at point $P$ to axle center at point $Q$ as illustrated in Fig.~\ref{fig:spring_lengths}(b) and Fig.~\ref{fig:spring_lengths}(c), by applying the cosine rule:
\begin{subequations}\label{eq:4}
	\begin{align}
		\gamma_{\mathrm{l}}&=\arccos \left(\frac{ d^2 + R_{\mathrm{h}}^2 - {l_{\mathrm{l,c}2\mathrm{c}}}^2 }{2   d   R_{\mathrm{h}}}\right)\label{eq:4A},\\
		\gamma_{\mathrm{r}} &=\arccos \left(\frac{ d^2 + R_{\mathrm{h}}^2 -{l_{\mathrm{r,c}2\mathrm{c}}}^2}{2   d   R_{\mathrm{h}}}\right) ,\label{eq:4B}
	\end{align}
\end{subequations}
where $R_{\mathrm{h}}$ represents the constant rack arm length from the origin $O$ to the spring attachment points $P_l$ and $P_r$, and where we can write $l_{\mathrm{l,c}2\mathrm{c}}$ and $l_{\mathrm{r,c}2\mathrm{c}}$ in terms of the spring elongations: 
\begin{align}
    l_{\mathrm{l,c}2\mathrm{c}} = \Delta l_{\mathrm{l}} - r_{\mathrm{Q}} - r_{\mathrm{P}} + l_0 & \\
    l_{\mathrm{r,c}2\mathrm{c}} = \Delta l_{\mathrm{r}} - r_{\mathrm{Q}} - r_{\mathrm{P}} + l_0.
\end{align}
Here, $r_{\mathrm{Q}}$ and $r_{\mathrm{P}}$ denote the radii of the pins on the motor crank and the rack, respectively (see Figs.~\ref{fig:spring_lengths}); 
$l_0$ is the spring rest length. 

The angle $\varphi_\mathrm{a}$ can be calculated using the angle $\gamma$ between $d$ and the line $\overline{OM}$. This calculation can be performed based on either the left spring length, where $\gamma_{\mathrm{l}} -  \gamma = \beta - \varphi_{\mathrm{a}}$, or from the right spring length, where $\gamma_{\mathrm{r}} +  \gamma = \beta + \varphi_{\mathrm{a}}$, where the constant $\beta$ is the angle between $R_{\mathrm{h}}$ and the symmetry line. We use the average of the two:
\begin{equation}
    \varphi_{\mathrm{a}} = (\gamma_\mathrm{r}-\gamma_\mathrm{l})/2 + \gamma.
\end{equation}
We can then express $\varphi_{\mathrm{a}}$ as a function of $\alpha$, $\Delta l_{\mathrm{l}}$ and $\Delta l_{\mathrm{r}}$, such that: 
\begin{align}
    \label{eq:phi_a_final}
    \varphi_{\mathrm{a}} =  \varphi_{\mathrm{a}}(\alpha, \Delta l_{\mathrm{l}}, \Delta l_{\mathrm{r}}) &= \frac{1}{2} \left( \arccos \left(\frac{ d^2 + R_{\mathrm{h}}^2 - {(\Delta l_{\mathrm{r}} -r_Q - r_P + l_0)}^2 }{2   d   R_{\mathrm{h}}}\right)\right) \nonumber\\ & - \frac{1}{2}  \left(  \arccos \left(\frac{ d^2 + R_{\mathrm{h}}^2 -{(\Delta l_{\mathrm{l}} -r_Q - r_P + l_0)}^2}{2   d   R_{\mathrm{h}}}\right)\right) \nonumber \\ & + \arcsin   \left(\frac{r}{d}\, \,\sin(\pi-\alpha) \right).  
\end{align}
\section{Torque Sensing}
\label{sec:torque_sensing_and_inverse_kinematics}
Our control architecture relies on accurate torque sensing to control the lean torque at the bicycle rack, $\tau_{\mathrm{a}}$, and to close the inner control loop on the crank torque, $\tau_{\mathrm{sc}}$. As we do not use dedicated torque sensors on the bicycle, we determine $\tau_{\mathrm{a}}$ from the forces in the left and right springs, which we reconstruct from the measured spring elongations and the mechanism geometry. Fig.~\ref{fig:fbd} shows the Free-Body diagrams (FBDs) of SPINEA and the definitions of the directions of all torques and forces.

We denote the forces that the left and right springs exert on the crank and the bicycle rack in the $\mathcal{N}$ triad by $\prescript{\mathcal{N}}{}{\V{F}_{\mathrm{l}}}$ and $\prescript{\mathcal{N}}{}{\V{F}_{\mathrm{r}}}$, respectively (Fig.~\ref{fig:fbd}a,c). We begin by expressing these forces in terms of their magnitudes and directions (Figs.~\ref{fig:spring_lengths} for kinematics and Fig.~\ref{fig:fbd} for the FBDs).

The left spring force is given by:
\begin{equation}
\label{eq:NFL}
		\prescript{\mathcal{N}}{}{\V{F}_{\mathrm{l}}}= F_{\mathrm{l}}   \hat{\V{e}}_{{Q/ P_\mathrm{l}}}  =  F_{\mathrm{l}}   \prescript{\mathcal{N}}{}{\V{r}_{{Q/P_\mathrm{l}}}} / \lvert \prescript{\mathcal{N}}{}{\V{r}_{Q/{P_\mathrm{l}}}} \rvert.
\end{equation}
and the right spring force is:
\begin{equation}
\label{eq:NFR}
		\prescript{\mathcal{N}}{}{\V{F}_{\mathrm{r}}}= F_{\mathrm{r}} \hat{\V{e}}_{{Q/P_\mathrm{r}}}  =   F_{\mathrm{r}}   \prescript{\mathcal{N}}{}{\V{r}_{{Q/P_\mathrm{r}}}} / \lvert \prescript{\mathcal{N}}{}{\V{r}_{{Q/P_\mathrm{r}}}} \rvert.      
\end{equation}
Here, ${F}_{\mathrm{l}}$ and ${F}_{\mathrm{r}}$ are the magnitudes of the left and right spring forces, respectively, with the unit vectors $\hat{\V{e}}_{{Q/P_\mathrm{r}}}$ and $\hat{\V{e}}_{{Q/P_\mathrm{r}}}$ pointing from $P_{\mathrm{l}}$ to $Q$ and from $P_{\mathrm{r}}$ to $Q$. The relative position vectors $\prescript{\mathcal{N}}{}{\V{r}_{{Q/P_\mathrm{l}}}}$ and $\prescript{\mathcal{N}}{}{\V{r}_{{Q/P_\mathrm{r}}}}$ represent the vector from point $P_\mathrm{l}$ to $Q$ and $P_\mathrm{r}$ to $Q$, respectively. 

At the rest length $l_0$, a pretensile force $F_0$ exists in the springs. Fig.~\ref{fig:Rest_length} illustrates the system configuration at this length. When the applied force exceeds the pretension, the springs elongate by $\Delta l_{\mathrm{l}}$ and $\Delta l_{\mathrm{r}}$, respectively. The resulting spring forces are proportional to the spring constants $k_1$ and $k_2$, and are computed using Hooke’s law: 
\begin{subequations}
    \begin{align}
	    F_{\mathrm{l}}  & =  k_{\mathrm{l}}   \Delta l _\mathrm{l}+ F_0  \label{eq: Sensing Forces left}\\& 
        = k_{\mathrm{l}}  (l_{\mathrm{l,c}2\mathrm{c}} +r_{\mathrm{Q}}+ r_{\mathrm{P}}-l_0)+ F_0\,, \quad \text{with} \quad l_{\mathrm{l,c}2\mathrm{c}}= \lvert\V{r}_{{Q/P_\mathrm{l}}}\rvert 
        \label{eq:Inverse_kinematics_left}
	\end{align}
	\begin{align}
		F_{\mathrm{r}} & =  k_{\mathrm{r}} \Delta l_\mathrm{r}+ F_0 \label{eq: Sensing Forces right} \\&   = 
        k_{\mathrm{r}} ( l_{\mathrm{r,c}2\mathrm{c}} +r_{\mathrm{Q}}+ r_{\mathrm{P}}-l_0)+ F_0\,, \quad \text{with} \quad  
       l_{\mathrm{r,c}2\mathrm{c}}= \lvert\V{r}_{{Q/P_\mathrm{r}}}\rvert \label{eq: Inverse kinematics right}
	\end{align}
    \label{eq:Springs}
\end{subequations}
For the calculation of the spring forces, we assume $k=k_{\mathrm{l}}=k_{\mathrm{r}}$, as both springs are identical and sourced with the same specifications.  

We use vector addition to calculate $\prescript{\mathcal{N}}{}{\V{r}_{{Q/P_\mathrm{l}}}}$ and $\prescript{\mathcal{N}}{}{\V{r}_{{Q/P_\mathrm{r}}}}$ as: 
\begin{align} \label{eq: position Pl to Q} \prescript{\mathcal{N}}{}{\V{r}_{{Q/P_{\mathrm{l}}}}}& = \prescript{\mathcal{N}}{}{\V{r}_{Q/O}} - \prescript{\mathcal{N}}{}{{C}_{{\mathcal{L}}}} \prescript{\mathcal{L}}{}{\V{r}_{{P_{\mathrm{l}}}/O}}  =  \prescript{\mathcal{N}}{}{\V{r}_{Q/O}} - \begin{pmatrix} 1 & 0 & 0 \\ 0 & \cos(\varphi_{\mathrm{a}}) & -\sin(\varphi_{\mathrm{a}}) \\ 0 & \sin(\varphi_{\mathrm{a}}) & \cos(\varphi_{\mathrm{a}}) \end{pmatrix}\prescript{\mathcal{L}}{}{\V{r}_{{P_{\mathrm{l}}/O}}}, \\
\label{eq: position Pr to Q}   \prescript{\mathcal{N}}{}{\V{r}_{{Q/P_{\mathrm{r}}}}} &= \prescript{\mathcal{N}}{}{\V{r}_{Q/O}} -\prescript{\mathcal{N}}{}{{C}_{{\mathcal{L}}}} \prescript{\mathcal{L}}{}{\V{r}_{{P_{\mathrm{r}}/O}}} = \prescript{\mathcal{N}}{}{\V{r}_{Q/O}} - \begin{pmatrix} 1 & 0 & 0 \\ 0 & \cos(\varphi_{\mathrm{a}}) & -\sin(\varphi_{\mathrm{a}}) \\ 0 & \sin(\varphi_{\mathrm{a}}) & \cos(\varphi_{\mathrm{a}}) \end{pmatrix} \prescript{\mathcal{L}}{}{\V{r}_{{P_{\mathrm{r}}/O}}} , \end{align}
with $\prescript{\mathcal{L}}{}{\V{r}_{{P_{\mathrm{l}}/O}}} = (0, - R_{\mathrm{h}}\sin(\beta), - R_{\mathrm{h}}\cos(\beta))^{\mathrm{T}}$ and $\prescript{\mathcal{L}}{}{\V{r}_{{P_{\mathrm{r}}/O}}} = (0, R_{\mathrm{h}}\sin(\beta), - R_{\mathrm{h}}\cos(\beta))^{\mathrm{T}}$.
The relative position vector $\prescript{\mathcal{N}}{}{\V{r}_{Q/O}}$ of $Q$ with respect to $O$ also results from vector addition: 
\begin{equation}
    \prescript{\mathcal{N}}{}{\V{r}_{{Q/O}}} = \prescript{\mathcal{N}}{}{\V{r}_{{Q/M}}} + \prescript{\mathcal{N}}{}{\V{r}_{M/O}} ,
\end{equation} 
where the vector from $O$ to $M$ is $\prescript{\mathcal{N}}{}{\V{r}_{M/O}} = \begin{pmatrix} 0, 0, - h  \end{pmatrix} ^{\mathrm{T}}$ 
and vector $\prescript{\mathcal{N}}{}{\V{r}_{{Q/M}}}$ represents the position vector from point $M$ to $Q$ (Fig.~\ref{fig:fbd}a). To obtain $\prescript{\mathcal{N}}{}{\V{r}_{{Q/M}}}$, we use the position vector from point $M$ to $Q$ in the crank triad $\mathcal{M}$ as $\prescript{\mathcal{M}}{}{\V{r}_{{Q/M}}} = \begin{pmatrix} 0 , 0 ,-r \end{pmatrix}^{\mathrm{T}}$. Then, we transform this vector from $\mathcal{M}$ into $\mathcal{N}$ rotating it by $\alpha$, 
\begin{equation}
    \prescript{\mathcal{N}}{}{\V{r}_{{Q/M}}}=
    \prescript{\mathcal{N}}{}{\M{C}_{{\mathcal{M}}}}
    \prescript{\mathcal{M}}{}{\V{r}_{{Q/M}}} = \begin{pmatrix} 1 & 0 & 0 \\ 0 & \cos(\alpha) & -\sin(\alpha) \\ 0 & \sin(\alpha) & \cos(\alpha) \end{pmatrix} \prescript{\mathcal{M}}{}{\V{r}_{{Q/M}}},
\end{equation}
with the rotation matrix $\prescript{\mathcal{N}}{}{\M{C}_{{\mathcal{M}}}}$.

Next, we can calculate the torque that the springs exert on the bicycle, $\V{\tau}_{\mathrm{sc}}=(\tau_{\mathrm{sc}},0,0)^\mathrm{T}$, in the $\mathcal{N}$ triad as: 
\begin{equation}
    \label{eq:crank_torque_sensing}
 	\prescript{\mathcal{N}}{}{\V{\tau}_{\mathrm{sc}}} = \prescript{\mathcal{N}}{}{\V{r}_{{Q/M}}} \times (\prescript{\mathcal{N}}{}{\V{F}_{\mathrm{l}}} + \prescript{\mathcal{N}}{}{\V{F}_{\mathrm{r}}}),
\end{equation}
where $\prescript{\mathcal{N}}{}{\V{F}_{\mathrm{l}}}$ and $\prescript{\mathcal{N}}{}{\V{F}_{\mathrm{r}}}$ are the spring forces as in (\ref{eq:NFL}) and (\ref{eq:NFR}). 

Substituting the magnitude of the spring forces (\ref{eq: Sensing Forces left}) and (\ref{eq: Sensing Forces right}), we obtain: 

\begin{align}
    \label{eq:crank_torque_sensing_expanded}
 	\prescript{\mathcal{N}}{}{\V{\tau}_{\mathrm{sc}}} = \prescript{\mathcal{N}}{}{\V{r}_{{Q/M}}} \times \left((k   \Delta l _\mathrm{l}+ F_0)\,\dfrac{\prescript{\mathcal{N}}{}{\V{r}_{{Q/P_\mathrm{l}}}} }{\lvert \prescript{\mathcal{N}}{}{\V{r}_{{Q/P_\mathrm{l}}}} \rvert} + (k   \Delta l _\mathrm{r}+ F_0)\,\dfrac{\prescript{\mathcal{N}}{}{\V{r}_{{Q/P_\mathrm{r}}}} }{\lvert \prescript{\mathcal{N}}{}{\V{r}_{{Q/P_\mathrm{r}}}} \rvert} \right)\,, \\
    \text{with} \quad \prescript{\mathcal{N}}{}{\V{r}_{{Q/M}}} = \begin{pmatrix} 0 , r\sin(\alpha) ,-r\cos(\alpha) \end{pmatrix}^{\mathrm{T}}\\
    \text{and} \quad \prescript{\mathcal{N}}{}{\V{r}_{{Q/P_{\mathrm{l}}}}} = \begin{pmatrix} 0 ,  r\sin(\alpha)  - R_{\mathrm{h}}\sin(\varphi_{\mathbf{a}} - \beta) ,-h -r\cos(\alpha) + R_{\mathrm{h}}\cos(\varphi_{\mathbf{a}} - \beta)  \end{pmatrix}^{\mathrm{T}} \\
    \text{and} \quad \prescript{\mathcal{N}}{}{\V{r}_{{Q/P_{\mathrm{r}}}}} = \begin{pmatrix} 0 ,r\sin(\alpha) - R_{\mathrm{h}}\sin(\varphi_{\mathbf{a}} + \beta) ,-h-r\cos(\alpha)+R_{\mathrm{h}}\cos(\varphi_{\mathbf{a}} + \beta) \end{pmatrix}^{\mathrm{T}}.
\end{align}

Finally, we can write ${\tau}_{\mathrm{sc}}$ in terms of our measured inputs $\alpha$, $\Delta l_{\mathrm{l}}$ and $\Delta l_{\mathrm{r}}$ and $\varphi_{\mathrm{a}}$ as in~(\ref{eq:phi_a_final}):
\begin{align}
    \label{eq:tau_sc_final}
    \tau_{\mathrm{sc}} = \tau_{\mathrm{sc}} (\alpha, \Delta l_{\mathrm{l}}, \Delta l_{\mathrm{r}}) &\nonumber=  r\sin(\alpha)(-h - r\cos(\alpha) + R_{\mathrm{h}} \cos(\varphi_{\mathrm{a}}(\alpha, \Delta l_{\mathrm{l}}, \Delta l_{\mathrm{r}}) - \beta))
    \frac{k \Delta l_{\mathrm{l}} + F_0}{\lvert\prescript{\mathcal{N}}{}{\V{r}_{{Q/P_\mathrm{l}}}} \rvert}\\ 
    &\nonumber+ r\cos(\alpha)(r\sin(\alpha) -  R_{\mathrm{h}}  \sin(\varphi_{\mathrm{a}}(\alpha, \Delta l_{\mathrm{l}}, \Delta l_{\mathrm{r}}) - \beta)) \frac{k\Delta l_{\mathrm{l}}+F_0}{\lvert\prescript{\mathcal{N}}{}{\V{r}_{{Q/P_\mathrm{l}}}} \rvert} \nonumber \\
    & \nonumber\quad +  r\sin(\alpha)(-h - r\cos(\alpha) + R_{\mathrm{h}} \cos(\varphi_{\mathrm{a}}(\alpha, \Delta l_{\mathrm{l}}, \Delta l_{\mathrm{r}}) + \beta)) \frac{k \Delta l_{\mathrm{r}} + F_0}{\lvert\prescript{\mathcal{N}}{}{\V{r}_{{Q/P_\mathrm{r}}}} \rvert}    
    \\&+ 
     r\cos(\alpha)( r\sin(\alpha) - R_{\mathrm{h}} \sin(\varphi_{\mathrm{a}}(\alpha, \Delta l_{\mathrm{l}}, \Delta l_{\mathrm{r}}) + \beta)) \frac{k \Delta l_{\mathrm{r}} + F_0}{\lvert\prescript{\mathcal{N}}{}{\V{r}_{{Q/P_\mathrm{r}}}} \rvert}. \\&
    \text{with}  \quad  \lvert\prescript{\mathcal{N}}{}{\V{r}_{{Q/P_\mathrm{l}}}} \rvert =  \sqrt{(r\sin(\alpha)  - R_{\mathrm{h}}\sin(\varphi_{\mathbf{a}} - \beta))^2 + (-h -r\cos(\alpha) + R_{\mathrm{h}}\cos(\varphi_{\mathbf{a}} - \beta))^2}, \\&
    \text{and}  \quad  \lvert\prescript{\mathcal{N}}{}{\V{r}_{{Q/P_\mathrm{r}}}} \rvert =  \sqrt{(  r\sin(\alpha)  - R_{\mathrm{h}}\sin(\varphi_{\mathbf{a}} + \beta))^2 + (-h -r\cos(\alpha) + R_{\mathrm{h}}\cos(\varphi_{\mathbf{a}} + \beta))^2 }  
\end{align}
Similarly to $\tau_{\mathrm{sc}}$, we can determine the torque $\tau_{\mathrm{a}}$ at the bicycle rack in terms of the spring forces $\prescript{\mathcal{N}}{}{\V{F}_{\mathrm{l}}}$ and $\prescript{\mathcal{N}}{}{\V{F}_{\mathrm{r}}}$:
\begin{equation}
    \label{eq:bike_torque_sensing}
 	\prescript{\mathcal{N}}{}{\V{\tau}_{\mathrm{a}}} = \prescript{\mathcal{N}}{}{\V{r}_{{P_{\mathrm{l}}/O}}} \times \prescript{\mathcal{N}}{}{\V{F}_{\mathrm{l}}}  + \prescript{\mathcal{N}}{}{\V{r}_{{P_{\mathrm{r}}/O}}} \times\prescript{\mathcal{N}}{}{\V{F}_{\mathrm{r}}}\,,
\end{equation}
which can also be expressed in scalar form in terms of our measured inputs $\alpha$, $\Delta l_{\mathrm{l}}$ and $\Delta l_{\mathrm{r}}$ and $\varphi_{\mathrm{a}}$ as in~(\ref{eq:phi_a_final}):

\begin{align}
    \label{eq:tau_a_final}
    \tau_{\mathrm{a}} = \tau_{\mathrm{a}}(\alpha, \Delta l_{\mathrm{l}}, \Delta l_{\mathrm{r}})  &=   R_{\mathrm{h}}\sin(\varphi_{\mathrm{a}}(\alpha, \Delta l_{\mathrm{l}}, \Delta l_{\mathrm{r}})- \beta)(-h - r\cos(\alpha) + R_{\mathrm{h}} \cos(\varphi_{\mathrm{a}}(\alpha, \Delta l_{\mathrm{l}}, \Delta l_{\mathrm{r}}) - \beta)) \dfrac{k \Delta l_{\mathrm{l}} + F_0}{\lvert\prescript{\mathcal{N}}{}{\V{r}_{{Q/P_\mathrm{l}}}} \rvert} \nonumber\\&+ R_{\mathrm{h}}\cos(\varphi_{\mathrm{a}}(\alpha, \Delta l_{\mathrm{l}}, \Delta l_{\mathrm{r}})- \beta)(r\sin(\alpha) - R_{\mathrm{h}} \sin(\varphi_{\mathrm{a}}(\alpha, \Delta l_{\mathrm{l}}, \Delta l_{\mathrm{r}}) - \beta)) \dfrac{k \Delta l_{\mathrm{l}} + F_0}{\lvert\prescript{\mathcal{N}}{}{\V{r}_{{Q/P_\mathrm{l}}}} \rvert}  \nonumber\\&
    + R_{\mathrm{h}}\sin(\varphi_{\mathrm{a}}(\alpha, \Delta l_{\mathrm{l}}, \Delta l_{\mathrm{r}})+ \beta)(-h - r\cos(\alpha) + R_{\mathrm{h}} \cos(\varphi_{\mathrm{a}}(\alpha, \Delta l_{\mathrm{l}}, \Delta l_{\mathrm{r}}) + \beta)) \dfrac{k \Delta l_{\mathrm{r}} + F_0}{\lvert\prescript{\mathcal{N}}{}{\V{r}_{{Q/P_\mathrm{r}}}} \rvert} \nonumber\\
    & + R_{\mathrm{h}}\cos(\varphi_{\mathrm{a}}(\alpha, \Delta l_{\mathrm{l}}, \Delta l_{\mathrm{r}})+ \beta)(r\sin(\alpha) - R_{\mathrm{h}} \sin(\varphi_{\mathrm{a}}(\alpha, \Delta l_{\mathrm{l}}, \Delta l_{\mathrm{r}}) + \beta)) \dfrac{k \Delta l_{\mathrm{r}} + F_0}{\lvert\prescript{\mathcal{N}}{}{\V{r}_{{Q/P_\mathrm{r}}}} \rvert}.
\end{align}

\section{Reference Mapping Assuming Steady State}
\label{sec:Inverse_kinematics}
\begin{figure*}
	\centering
	\includegraphics[width=0.8\textwidth]{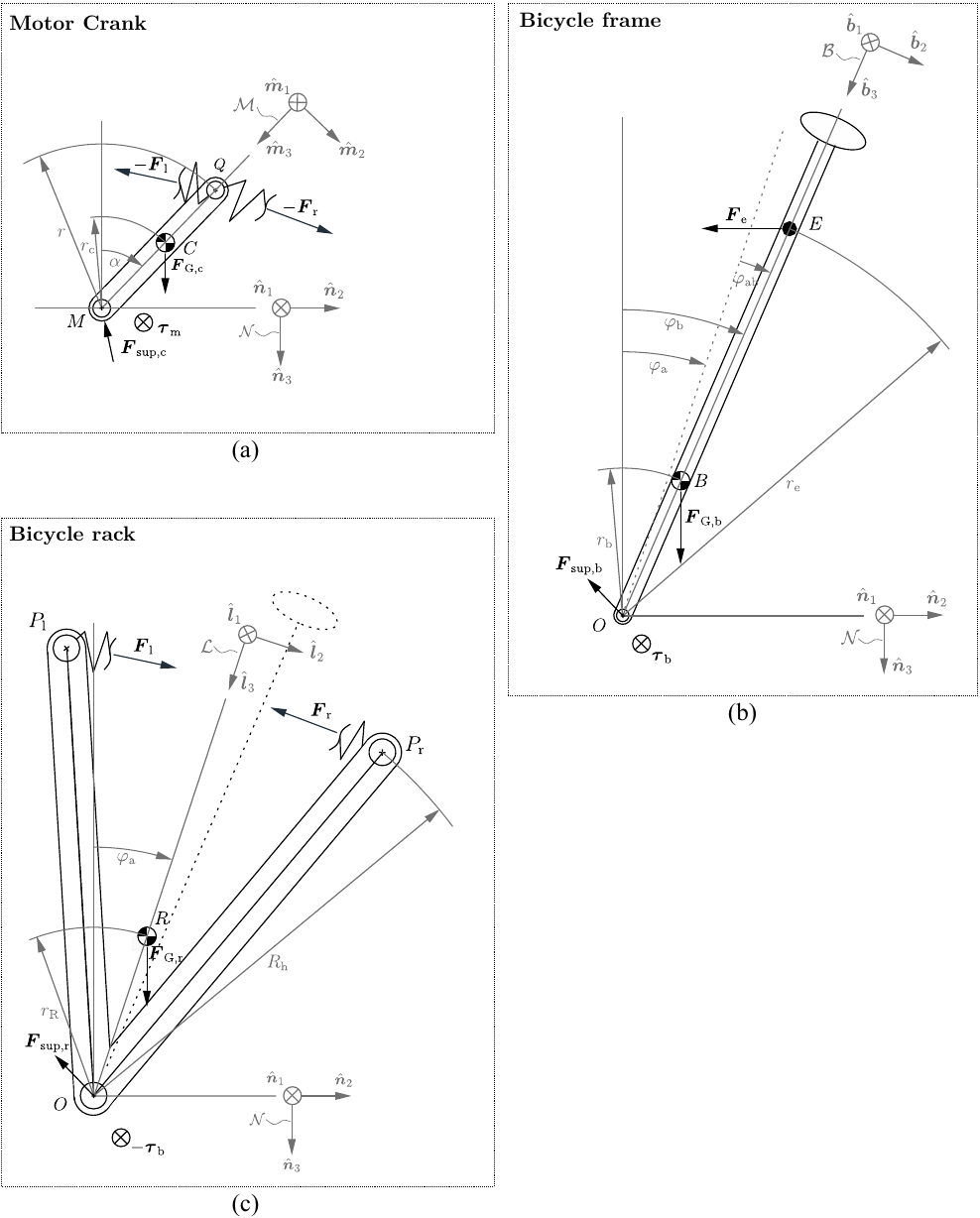}
	\caption{Free-Body diagrams (FBDs) of SPINEA and bicycle components, distinguishing crank, rack, and bicycle frame. (a) FBD of the motor crank. Cutting the springs on either side of the crank exposes the left spring force $\V{F}_{\mathrm{l}}$ and the right spring force $\V{F}_{\mathrm{r}}$. Cutting at the bearing of the crank introduces the support force vector $\V{F}_{\mathrm{sup,c}}$. The gravitational force $\V{F}_{\mathrm{G,c}}$ of the motor crank acts at the center of mass (CoM) $C$. The crank is actuated by a motor torque $\V{\tau}_{\mathrm{m}}$. The crank-fixed triad $\mathcal{M}$ rotates with respect to the inertial reference triad $\mathcal{N}$. (b) FBD of the bicycle frame. The external excitation motor, as used in the impedance identification experiments, exerts the force $\V{F}_{\mathrm{e}}$ on the attachment point $E$. Cutting at the bearing of the bicycle frame exposes the supporting force $\V{F}_{\mathrm{sup, b}}$. The bending torque $\V{\tau}_{\mathrm{b}}$ is a lumped reaction to a deviation $\varphi_{\mathrm{ab}}$ between the lean angle $\varphi_{\mathrm{a}}$ of the bicycle rack and the lean angle $\varphi_{\mathrm{b}}$ of the bicycle frame. The frame's gravitational force $\V{F}_{\mathrm{G, b}}$ acts at the frame's CoM $B$. The triad $\mathcal{B}$ is rotating with the bicycle frame. (c) FBD of the bicycle rack. Cutting the springs on either side of the rack exposes forces $\V{F}_{\mathrm{l}}$ at point $P_{\mathrm{l}}$ and $F_{\mathrm{r}}$ at point $P_{\mathrm{r}}$. The bearing of the bicycle rack exerts a supporting force $\V{F}_{\mathrm{sup, r}}$. The gravitational force $\V{F}_\mathrm{G,r}$ acts on the bicycle rack in its CoM $R$. The reaction torque that the bicycle frame exerts on the rack is $-\V{\tau}_{\mathrm{b}}$. The rack-fixed triad $\mathcal{L}$ rotates with respect to $\mathcal{N}$.}
	\label{fig:fbd}
\end{figure*}

The control scheme requires mapping a reference torque $\tau_{\mathrm{a, \mathrm{ref}}}$ (given by either the bicycle model or an external torque reference) to a torque reference at crank level, $\tau_{\mathrm{sc, \mathrm{ref}}} $. However, in principle there is no one-to-one mapping given the additional degree of freedom in the series elastic feature of SPINEA. Therefore, an additional assumption is necessary. We choose to assume that the system has already converged to steady state, where crank acceleration is zero, which enables solving the equations. 
With this assumption, the equations of motion of the drive unit and crank (Fig.~\ref{fig:fbd}a),
\begin{equation}
 	\prescript{\mathcal{N}}{}J_{\mathrm{du}\,\mathcal{N}} \prescript{\mathcal{N}}{}{ \ddot{\V{\alpha}}}= \prescript{\mathcal{N}}{}{\V{r}_{{Q/M}}} \times (-\prescript{\mathcal{N}}{}{\V{F}_{\mathrm{l}}} -\prescript{\mathcal{N}}{}{\V{F}_{\mathrm{r}}}) + \prescript{\mathcal{N}}{}{\V{\tau}_{\mathrm{m}}}+ \prescript{\mathcal{N}}{}{\V{\tau}_{\mathrm{G,c}}} \quad, 
    \label{eq:EOM_crank}
\end{equation}
can be simplified. In this equation, $\prescript{\mathcal{N}}{}J_{\mathrm{du}\,\mathcal{N}}$ is the mass moment of inertia tensor of the drive unit, including the crank, and $\prescript{\mathcal{N}}{}{ \ddot{\V{\alpha}}} = \begin{pmatrix} \ddot{\alpha}, 0 , 0 \end{pmatrix} $ is the angular crank acceleration. The term $\prescript{\mathcal{N}}{}{\V{\tau}_{\mathrm{m}}} = \begin{pmatrix} \tau_{\mathrm{m}}, 0 , 0 \end{pmatrix}^{\mathrm{T}}$ represents the torque applied by the drive unit. The torque resulting from gravity acting on the crank is given by $\prescript{\mathcal{N}}{}{\V{\tau}_{\mathrm{G,c}}}$. 

Assuming that the crank is massless, we neglect both the gravitational term ($\prescript{\mathcal{N}}{}{\V{\tau}_{\mathrm{G,c}}}$) and the inertial term ($\prescript{\mathcal{N}}{}J_{\mathrm{du}\,\mathcal{N}} \prescript{\mathcal{N}}{}{ \ddot{\V{\alpha}}}$) in (\ref{eq:EOM_crank}), such that $\tau_\mathrm{m}=\tau_\mathrm{sc}$, i.e.\ the motor applies a torque that is equal and opposite to the torque applied on the crank by the springs. Note that in the definitions above, while $\tau_\mathrm{sc}$ was defined positive in counterclockwise direction, its ``reaction'' $\tau_\mathrm{m}$ was defined positive in clockwise direction. This was done to maintain consistent signs in the control design.

The assumption of no mass means that the crank moves instantaneously, and its angle $\alpha$ results from solving the system of equations given by (\ref{eq:Inverse_kinematics_left}), (\ref{eq: Inverse kinematics right}), (\ref{eq: position Pl to Q}), (\ref{eq: position Pr to Q}), and (\ref{eq:tau_a_final}) for given $\tau_{\mathrm{a}}=\tau_{\mathrm{a, \mathrm{ref}}}$ and the angle $\varphi_a$ equal to the currently measured angle.

We did not find a closed analytic solution for $\alpha_{\mathrm{ref}}$ from the system of equations, such that we solve it iteratively. We find $\alpha_{\mathrm{ref}}$ via an iterative bisection interpolation method performed over 20 iterations. This method involves iteratively adjusting the bisectional crank angle $\alpha$ until $\tau_{\mathrm{a}}$ matches $\tau_{\mathrm{a,ref}}$ (i.e., the reference input).

Specifically, each iteration involves the following three steps:

\begin{itemize}
\item 
In step 1, we use the geometric relationships to determine the spring elongations as functions of the crank angle $\alpha_{\mathrm{ref}}$ (the variable that will be updated iteratively) and the current rack lean angle $\varphi_{\mathrm{a}}$ (the measured input):
\begin{align}
    \Delta l_{\mathrm{l}}(\alpha_{\mathrm{ref}}, \varphi_{\mathrm{a}}) = & \quad l_{\mathrm{l,c}2\mathrm{c}} +r_{\mathrm{Q}}+ r_{\mathrm{P}}-l_0 \\& =  \sqrt{(-h - r\cos(\alpha_{\mathrm{ref}}) + R_{\mathrm{h}} \cos(\varphi_{\mathrm{a}} - \beta))^2 
    + (r\sin(\alpha_{\mathrm{ref}}) - R_{\mathrm{h}} \sin(\varphi_{\mathrm{a}} - \beta))^2} +r_{\mathrm{Q}}+ r_{\mathrm{P}}-l_0 \\
    \Delta l_{\mathrm{r}}(\alpha_{\mathrm{ref}},\varphi_{\mathrm{a}}) = &  \quad l_{\mathrm{r,c}2\mathrm{c}} +r_{\mathrm{Q}}+ r_{\mathrm{P}}-l_0 \\& = \sqrt{(-h - r\cos(\alpha_{\mathrm{ref}}) + R_{\mathrm{h}} \sin(\varphi_{\mathrm{a}} + \beta))^2 
    + (r\sin(\alpha_{\mathrm{ref}}) - R_{\mathrm{h}} \sin(\varphi_{\mathrm{a}} + \beta))^2} +r_{\mathrm{Q}}+ r_{\mathrm{P}}-l_0
\end{align}

\item In step 2, we substitute the spring elongations calculated in step 1, along with $\alpha_{\mathrm{ref}}$ and the currently measured $\varphi_{\mathrm{a}}$ into equation (\ref{eq:tau_a_final}) to determine the torque $\tau_{\mathrm{a, \mathrm{ref}}}$.

\item In step 3, we calculate the error in $\tau_{\mathrm{a, \mathrm{ref}}}$ and correct $\alpha_{\mathrm{ref}}$ according to the interpolation algorithm.

\end{itemize}
To initialize the algorithm, we use the value of $\alpha_{\mathrm{ref}}$ from the previous time step.
After the number of iterations has been completed, we set the reference angle $\alpha_{\mathrm{ref}}$ to the final $\alpha$, and we calculate $\tau_{\mathrm{sc,ref}}$ from (\ref{eq:tau_sc_final}) for this final value of $\alpha$ and the current rack lean angle $\varphi_{\mathrm{a}}$.
\begin{table}
\centering
    \caption{Design parameters of four-bar mechanism, SPINEA and excitation mechanism}  \label{tab:Constant_parameters_supplementary}
    \centering
    \begin{tabular}{p{5cm} p{4cm}}
        \hline
        Parameter & Value \\
        \hline
        \textit{Four-bar mechanism} & \\
        Radius between point $H$ and point $O$ &  $R_{\mathrm{h}} = \SI{0.6}{\meter}$ \\
        Length of the fixed link &  $b = \SI{0.42}{\meter}$ \\
        Length of coupler &  $c = \SI{0.184}{\meter}$ \\
        Radius of driven linkage (crank) &  $r = \SI{0.239}{\meter}$ \\
        Angle between the base linkage and the horizontal axis &  $\upsilon = \SI{1.571}{\radian}$ \\
        Offset angle between bicycle's symmetry axis and connection point $H$ &  $\zeta = \SI{-0.25}{\radian}$ \\
        Saddle bag height & $h_{\mathrm{bag}} = \SI{0.45}{\meter}$\\
        Saddle bag width & $w_{\mathrm{bag}} = \SI{0.4}{\meter}$\\
        Height of bicycle rack & $h_{\mathrm{rack}} = \SI{0.74}{\meter}$\\
        \textit{SPINEA} & \\
        Height of motor crank axis above lean axis & $h = \SI{0.492}{\meter}$  \\
        Radius of motor crank & $r = \SI{0.17}{\meter}$  \\
        Radius bike ground hinge to spring attachment at bicycle rack & $R_{\mathrm{h}} = \SI{0.685}{\meter}$   \\
        Angle between spring attachment on bicycle rack and symmetry line  & $\beta = \SI{0.258}{\radian}$ \\
        Radius at $P_{\mathrm{l}}$ and $P_{\mathrm{r}}$& $r_{P} = \SI{0.008}{\meter}$   \\
        Radius of pin at $Q$ & $r_{Q} = \SI{0.006}{\meter}$   \\
        Spring constant of coil springs  & $k = k_1 = k_2 = \SI{8360}{\newton\meter^{-1}}$ \\
        Preloaded length of coil springs & $l_\mathrm{0} = \SI{0.154}{\meter}$ \\
        Pretension force of coil springs &  $F_0 = \SI{114.54}{\newton}$ \\
        \textit{Excitation mechanism} & \\
        Height of fixation of external excitation rod  & $h_{\mathrm{e}} = \SI{0.69}{\meter}$ \\
        Radius of excitation motor crank & $r_\mathrm{e} = \SI{0.1}{\meter}$   \\
        \hline
    \end{tabular}
\end{table}
\newpage
\section{Theoretical transfer functions for impedance and torque tracking}
\label{sec:Linearized transfer functions}
To assess the theoretical dynamic behaviour of SPINEA we derive the transfer function of the mechanical impedance $Z(s) := -\frac{\tau_{\mathrm{a}}(s)}{s\varphi_{\mathrm{a}}(s)}$ and the transfer function for torque tracking performance $G_{\mathrm{track}}(s) := \frac{\tau_{\mathrm{a}}(s)}{\tau_{\mathrm{a,ref}}(s)}$. These closed‑form linearized expressions describe the expected frequency domain response. In the main part of the paper we compare the analytical predictions with experimental measurements, identify any discrepancies, and draw conclusions about the system dynamics and controller performance.

We base the derivation on the equation of motion of the drive unit and crank (neglecting the gravitational moment): 
\begin{equation}
    J_{\mathrm{du}} \ddot{\alpha} = \tau_{\mathrm{m}} - \tau_{\mathrm{sc}}. 
    \label{eq:eom_imp}
\end{equation}

If we linearize $\tau_{\mathrm{sc}}$, we get: 
\begin{equation}
    \tau_{\mathrm{sc}} (\alpha, \varphi_{\mathrm{a}}) \approx k_{\mathrm{sc}, \alpha} \alpha  + k_{\mathrm{sc}, \varphi_{\mathrm{a}}}\varphi_{\mathrm{a}} ,  
\end{equation}
for this we need to compute the gradient with respect to $\alpha$ and $\varphi_{\mathrm{a}}$ at the operating point $(\alpha_0, \varphi_{\mathrm{a},0})$:
\begin{equation}
    \left[
    k_{\mathrm{a,\alpha}}
    \quad
    k_{\mathrm{a,\varphi_{\mathrm{a}}}}
    \right] = 
    \left[
    \left.\frac{\partial \tau_\mathrm{a}}{\partial \alpha}\right|_{(\alpha_0, \varphi_{\mathrm{a},0})}
    \quad
    \left.\frac{\partial \tau_\mathrm{a}}{\partial \varphi_\mathrm{a}}\right|_{(\alpha_0, \varphi_{\mathrm{a},0})}
    \right].
\end{equation}
and similarly we get for $\tau_{\mathrm{a}}$:
\begin{equation}
    \tau_{\mathrm{a}} (\alpha, \varphi_{\mathrm{a}}) \approx k_{\mathrm{a}, \alpha} \alpha + k_{\mathrm{a}, \varphi_{\mathrm{a}}}\varphi_{\mathrm{a}}, 
    \label{eq:tau_a_imp}
\end{equation}
with 
\begin{equation}
    \left[
    k_{\mathrm{sc,\alpha}}
    \quad
    k_{\mathrm{sc,\varphi_{\mathrm{a}}}}
    \right] = 
    \left[
    \left.\frac{\partial \tau_\mathrm{sc}}{\partial \alpha}\right|_{(\alpha_0, \varphi_{\mathrm{a},0})}
    \quad
    \left.\frac{\partial \tau_\mathrm{sc}}{\partial \varphi_\mathrm{a}}\right|_{(\alpha_0, \varphi_{\mathrm{a},0})}
    \right].
\end{equation}

Note that since the Hessian of the spring's potential with respect to the generalized coordinates $\alpha$ and $\varphi_\mathrm{a}$ is symmetric, we know that with the given definition of signs it must hold that $k_{\mathrm{sc,\varphi_a}}=-k_{\mathrm{a,\alpha}}$.

To determine $\alpha_{\mathrm{ref}}$, we use the linearized $\tau_{\mathrm{a}}$ and solve it:
\begin{equation}
    \alpha_{\mathrm{ref}} = \frac{1}{k_{\mathrm{a}, \alpha}} (\tau_{\mathrm{a, ref}} - k_{\mathrm{a}, \varphi_{\mathrm{a}}}\varphi_{\mathrm{a}}). 
\end{equation}
With this we can solve for $\tau_{\mathrm{sc, ref}}$
\begin{equation}
    \tau_{\mathrm{sc,ref}} = \frac{k_{\mathrm{sc}, \alpha}}{k_{\mathrm{a}, \alpha}} \tau_{\mathrm{a,ref}} + (k_{\mathrm{sc}, \varphi_{\mathrm{a}}}- k_{\mathrm{sc}, \alpha} \frac{k_{\mathrm{a},\varphi_{\mathrm{a}}}}{k_{\mathrm{a}, \alpha}}) \varphi_{\mathrm{a}}.
    \label{eq:tauscref}
\end{equation}

Closing the loop with the controller in steady-state, i.e.\ substituting~(\ref{eq_tau_ref_m}) from the main manuscript for $\tau_\mathrm{m}$ in~(\ref{eq:eom_imp}) gives: 
\begin{equation}
    (J_{\mathrm{du}}s^2 + Ks) \alpha = \left(1 + \left(P_{\tau} + \frac{I_{\tau}}{s} \right)\right) (\tau_\mathrm{sc, ref} - \tau_{\mathrm{sc}}).
    \label{eq:eom_with_control}
\end{equation}
By substituting $\tau_\mathrm{sc, ref}$ from \eqref{eq:tauscref} in \eqref{eq:eom_with_control}, we can solve for $\frac{\alpha(s)}{\varphi_{\mathrm{a}}(s)}$:

\begin{equation}
    \frac{\alpha(s)}{\varphi_{\mathrm{a}}(s)} = \frac{\left(1 + P_{\tau} + \frac{1}{I_{\tau}s}\right) \left(-k_{\mathrm{sc},\alpha} \frac{k_{\mathrm{a},\varphi_{\mathrm{a}}}}{k_{\mathrm{a}, \alpha}}\right)}{J_{\mathrm{du}}s^2 + Ks + \left(1 + P_{\tau} + \frac{I_{\tau}}{s}\right)k_{\mathrm{sc},\alpha}} \quad , 
\end{equation}

which we can use in~(\ref{eq:tau_a_imp}) to get the final impedance function: 
\begin{equation}
\begin{split}
Z(s) 
= \frac{-J_{\mathrm{du}}k_{\mathrm{a},\varphi_{\mathrm{a}}}s^2 - Kk_{\mathrm{a},\varphi_{\mathrm{a}}}s}{J_{\mathrm{du}}s^3 + Ks^2 + k_{\mathrm{sc}, \alpha} \left(1 + P_{\tau} \right)s + I_{\tau} k_{\mathrm{sc}, \alpha}}. 
\label{eq:Z_imp_theory}
\end{split}
\end{equation}

For the torque tracking we get the final formulation as: 
\begin{equation}
\begin{split}
    G_{\mathrm{track}}(s)  =   \frac{ k_{\mathrm{sc},\alpha} \left(1 + P_{\tau}\right) s + k_{\mathrm{sc},\alpha} I_{\tau} }{J_{\mathrm{du}}s^3 + Ks^2 +  k_{\mathrm{sc},\alpha} \left(1 + P_{\tau}\right) s + k_{\mathrm{sc},\alpha} I_{\tau}}.
\end{split}
\end{equation}

\section{Power Spectral Densities (PSDs) for Cycling Task}
\label{sec:PSD}
\begin{figure*}[h]
	\centering
	\includegraphics{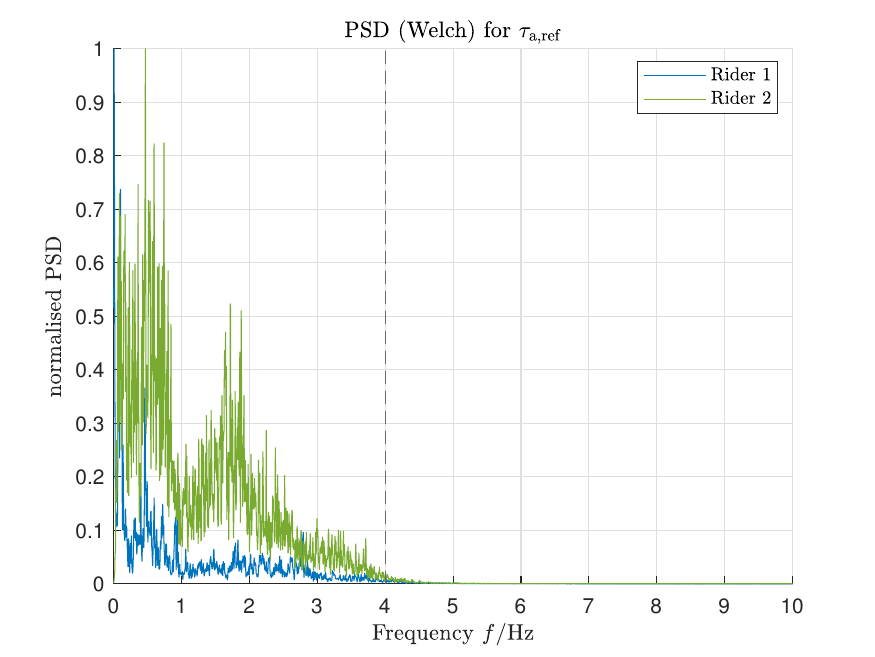}[hb]
	\caption{Power spectral density (PSD) for $\tau_{\mathrm{a, ref}}$ for two riders (Rider 1: blue, Rider 2: green) recorded during the cycling task on the simulator. PSDs were computed using Welch’s method. Vertical grey dashed lines indicates \SI{4}{\hertz}.}
	\label{fig:PSD}
\end{figure*}

To determine which frequency components are present in the reference torque $\tau_{\mathrm{a, ref}}$ we estimated its power‑spectral density (PSD) for the two test subjects. The PSDs, computed with Welch’s method, are shown in Fig.\ref{fig:PSD}. The bulk of the signal energy is concentrated below \SI{2}{\hertz}; nevertheless, the measurements contain measurable spectral content up to roughly \SI{4}{\hertz} (vertical grey dashed line).  

Because the torque‑tracking performance of SPINEA can only be meaningfully evaluated over the frequency range where the reference signal carries energy, the PSD analysis defines the upper limit for the tracking study. In the present data set, the relevant bandwidth therefore extends to about \SI{4}{\hertz}; beyond this frequency the reference torque would not provide useful information.
\section{Rationale for Bode Plots}
To validate that the system behaves linearly within its operational range for the Bode plot analysis, we conducted disturbance rejection experiments with an offset angle. Unlike the initial tests centered around the bicycle's upright configuration ($\varphi_{\mathrm{a},0} = \SI{0}{\degree}$), these experiments introduced non-zero initial bicycle positions ($\varphi_{\mathrm{a},0} \neq \SI{0}{\degree}$) to evaluate how the system responds when excited away from its neutral position. 

We kept the experimental setup identical to the previous trials, applied a concatenated sinusoidal signal whose maximum amplitude we scaled with frequency to perturb the system, and maintained a zero reference torque ($\tau_{\mathrm{a, ref}} = \SI{0}{\newton\meter}$). We investigated two different $\varphi_{\mathrm{a},0}$, namely $\varphi_{\mathrm{a},0}\approx\SI{-2}{\degree}$ and $\varphi_{\mathrm{a},0}\approx\SI{-4}{\degree}$. Because the excitation mechanism’s workspace is confined, setting the initial lean angle $\varphi_{\mathrm{a},0} \approx \SI{-4}{\degree}$ limited the the maximum amplitude to $\SI{3.5}{\degree}$. For the smaller initial angle ($\varphi_{\mathrm{a},0}\approx\SI{-2}{\degree}$) we conducted experiments with both the full amplitude of $\SI{7}{\degree}$ and the reduced amplitude of $\SI{3.5}{\degree}$ in order to assess the effect of the maximal excitation amplitude.

\begin{figure}
	\centering
	\includegraphics[width = 0.8\textwidth]{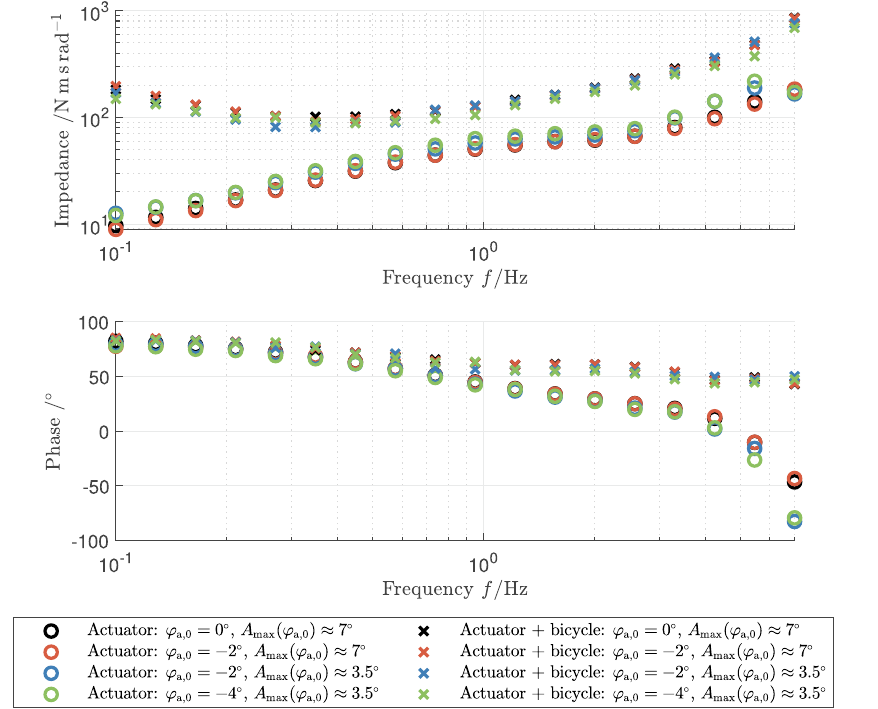}
	\caption{Impedance magnitude and phase versus frequency for the four experimental configurations. Colours encode the off‑center angle ($\varphi{\mathrm{a}} \neq \SI{0}{\degree}$) and maximal amplitude $A_{\mathrm{max}}$. Red markers correspond to
    $\varphi{\mathrm{a}} = \SI{-2}{\degree}$ with $A_{\mathrm{max}}(\varphi_{\mathrm{a},0}) = \SI{7}{\degree}$; blue markers represent $\varphi{\mathrm{a}} = \SI{-2}{\degree}$ with $A_{\mathrm{max}}(\varphi_{\mathrm{a},0}) = \SI{3.5}{\degree}$; and green markers indicate $\varphi{\mathrm{a}} = \SI{-4}{\degree}$ with $A_{\mathrm{max}}(\varphi_{\mathrm{a},0}) = \SI{3.5}{\degree}$.
    Black markers denote the neutral‑position case ($\varphi{\mathrm{a}} = \SI{0}{\degree}$).} 
	\label{fig:Impedance_Offset}
\end{figure}

In summary, we compare four experimental conditions:  
\begin{itemize}
  \item $\varphi_{\mathrm{a},0}=\SI{0}{\degree}$ with a maximum amplitude of $A_{\mathrm{max}}(\varphi_{\mathrm{a}}) = \SI{7}{\degree}$,
  \item $\varphi_{\mathrm{a},0}=\SI{-2}{\degree}$ with a maximum amplitude of $A_{\mathrm{max}}(\varphi_{\mathrm{a}}) = \SI{7}{\degree}$,
  \item $\varphi_{\mathrm{a},0}=\SI{-2}{\degree}$ with a maximum amplitude of $A_{\mathrm{max}}(\varphi_{\mathrm{a}}) = \SI{3.5}{\degree}$,
  \item $\varphi_{\mathrm{a},0}=\SI{-4}{\degree}$ with a maximum amplitude of $A_{\mathrm{max}}(\varphi_{\mathrm{a}}) = \SI{3.5}{\degree}$.
\end{itemize}
Fig.~\ref{fig:Impedance_Offset} presents the phase and magnitude responses derived from the offset experiments, compared with results obtained from the neutral position configuration. Results indicate that both phase and magnitude responses preserve their characteristic profiles across the tested range of frequencies, akin to those observed in tests conducted under the upright configuration. 

The consistency in the shape of the phase and magnitude responses suggests that the system's behaviour is sufficiently linear even when excited about a non-zero initial bicycle position. For $A_{\mathrm{max}}(\varphi_{\mathrm{a}}) = \SI{7}{\degree}$, the actuator impedance is unchanged when the bicycle is excited away from its neutral position. The impedance for $\varphi_{\mathrm{a},0}=\SI{0}{\degree}$ to $\varphi_{\mathrm{a},0}=\SI{-2}{\degree}$ overlap.

The actuator impedance remains unchanged for a maximum lean angle excitation of $A_{\mathrm{max}}(\varphi_{\mathrm{a}}) = \SI{7}{\degree}$ when the bicycle has a non-zero initial bicycle position; the curves for $\varphi_{\mathrm{a},0}=\SI{0}{\degree}$ to $\varphi_{\mathrm{a},0}=\SI{-2}{\degree}$ virtually coincide (see Fig.~\ref{fig:Impedance_Offset}). For the lower excitation level $A_{\mathrm{max}}(\varphi_{\mathrm{a}}) = \SI{3.5}{\degree}$, the impedance curves for $\varphi_{\mathrm{a},0}=\SI{-2}{\degree}$ and $\varphi_{\mathrm{a},0}=\SI{-4}{\degree}$ also overlap, but they are approximately 20\% higher than the corresponding curves obtained with $A_{\mathrm{max}}(\varphi_{\mathrm{a}}) = \SI{7}{\degree}$ over the entire frequency range. Nevertheless, this increase is small compared with the bicycle impedance.

This increase does not reflect a change in dynamic properties of the actuator. It is a consequence of static (Coulomb) friction. The friction torque is nearly constant, so its proportion relative to the applied torque grows when the excitation amplitude is decreased. Consequently, the ratio $\tau_{\mathrm{a}}/\dot\varphi_{\mathrm{a}}$, i.e. the identified impedance appears larger, although the underlying dynamics are unchanged.

The measured impedance responses for the combined actuator‑bicycle at $\varphi_{\mathrm{a},0}=\SI{0}{\degree}$, $\varphi_{\mathrm{a},0}=\SI{-2}{\degree}$, $\varphi_{\mathrm{a},0}=\SI{-4}{\degree}$ do not exhibit a systematic trend. 

Notably, no additional peaks or phase shifts were detected in the frequency response, which could indicate potential non-linearities within this range of motion. This supports our choice of Bode plots to analyse impedance.

\end{document}